\documentclass[journal]{IEEEtran}

\usepackage{xcolor,soul,framed} 

\colorlet{shadecolor}{yellow}
\usepackage[pdftex]{graphicx}
\graphicspath{{../pdf/}{../jpeg/}}
\DeclareGraphicsExtensions{.pdf,.jpeg,.png}

\usepackage[cmex10]{amsmath}
\usepackage{array}
\usepackage{mdwmath}
\usepackage{mdwtab}
\usepackage{eqparbox}
\usepackage{url}
\usepackage{amsmath}
\usepackage{amssymb} 
\usepackage{comment}
\usepackage{booktabs}
\usepackage{multirow}
\usepackage{siunitx}
\usepackage{adjustbox}
\usepackage{blindtext}
\usepackage{epsfig}
\usepackage[ruled,lined]{algorithm2e}
\usepackage[hidelinks]{hyperref}
\usepackage[capitalise]{cleveref}
\usepackage{soul}
\usepackage[numbers]{natbib}
\usepackage{pifont}
\usepackage{longtable}
\SetKwComment{Comment}{/* }{ */}

\newcommand{\mynorm}[1]{ \left\| #1 \right\| }
\usepackage{soul}

\usepackage{xspace}

\newcommand{\ie}{{\it i.e.},\xspace}

\begin{document}
\bstctlcite{IEEEexample:BSTcontrol}
\title{When Multi-Task Learning Meets Partial Supervision: A Computer Vision Review}
\author{
  Maxime Fontana, Michael Spratling, and Miaojing Shi,~\IEEEmembership{Senior Member,~IEEE}
  \thanks{Manuscript received June 23, 2023; revised March 14, 2024; accepted July 7, 2024. \textit{(Corresponding Author : Miaojing Shi)}}
  \thanks{{Maxime Fontana} is with the Department of Informatics, King's College London, London WC2B 4BG, United Kingdom (e-mail: maxime.fontana@kcl.ac.uk)} 
  \thanks{{Michael Spratling} is with the Department of Behavioural and Cognitive Sciences, University of Luxembourg, L-4366 Esch-sur-Alzette, Luxembourg and the Department of Informatics, King's College London, London WC2B 4BG, United Kingdom (e-mail: michael.spratling@uni.lu)} 
  \thanks{{Miaojing Shi} is with the College of Electronic and Information Engineering, Zip code : 201804, Tongji University, and with the Shanghai Institute of Intelligent Science and Technology, Tongji University (e-mail: mshi@tongji.edu.cn)} 
}

\markboth{M.Fontana et al.}%
{Shell \MakeLowercase{\textit{et al.}}: A Sample Article Using IEEEtran.cls for IEEE Journals}

\maketitle


\begin{abstract}
Multi-Task Learning (MTL) aims to learn multiple tasks simultaneously while exploiting their mutual relationships. By using shared resources to simultaneously calculate multiple outputs, this learning paradigm has the potential to have lower memory requirements and inference times compared to the traditional approach of using separate methods for each task. Previous work in MTL has mainly focused on fully-supervised methods, as task relationships can not only be leveraged to lower the level of data-dependency of those methods but they can also improve performance. However, MTL introduces a set of challenges due to a complex optimisation scheme and a higher labeling requirement. 
This review focuses on how MTL could be utilised under different partial supervision settings to address these challenges. First, this review analyses how MTL traditionally uses different parameter sharing techniques to transfer knowledge in between tasks. Second, it presents the different challenges arising from such a multi-objective optimisation scheme. Third, it introduces how task groupings can be achieved by analysing task relationships. Fourth, it focuses on how partially supervised methods applied to MTL can tackle the aforementioned challenges. Lastly, this review presents the available datasets, tools and benchmarking results of such methods. The reviewed papers, categorised following our work, are available: \href{https://github.com/Klodivio355/MTL-CV-Review}{https://github.com/Klodivio355/MTL-CV-Review}.
\end{abstract}


%

\IEEEpeerreviewmaketitle


\section{Introduction}

Convolutional Neural Networks (CNNs) have achieved great success in numerous and diverse computer vision tasks such as classification \cite{classification-sota, classification-success-1, classification-success-2, classifcation-success-3}, semantic segmentation \cite{semantic-segmentation-sota, Fast-R-CNN, unet, semantic-success1} and object-detection \cite{object-detection-sota, YOLOv3, semantic-success1}. These models have the common characteristic of being task specific. However, systems should ideally be capable of sharing knowledge between tasks.

Multi-Task Learning (MTL) \cite{Caruana} aims at providing computational models able to learn multiple tasks. To achieve this, MTL seeks to partition representations into task-agnostic and task-specific features so that each task can utilise a common representation. This is justified by previous work investigating the learning of representations in CNNs that distinguish two types of features. Firstly, shallow layers, which learn simple patterns (\ie edges and colors), are task-agnostic and should be shared. Secondly, deep layers which learn complex patterns (\ie objects), should be kept task-specific \cite{visualizing-CNN}. However, determining how to partition a specific network hierarchy is not trivial and depends on the tasks at hand \cite{cross-stich}. 
Nonetheless, MTL could help discover relationships and structure amongst tasks \cite{predictive-structures, task-relationships} which could improve performance compared to task-specific models. From a computational efficiency perspective, sharing representations results in enhanced memory efficiency and a significant reduction in inference time as shared representations only need to be inferred once to predict multiple tasks. 

Deep Learning (DL) models generally suffer from a high data-dependency during training, but acquiring large volumes of labeled data is not always feasible. This has motivated the development of various partial supervision configurations, with the unifying goal to create data-efficient DL solutions \cite{partly-supervised-MTL, MTPSL, semi-supervised-MTL-for-semantics-depth, Semi-supervised-MTL, semi-MTL-for-lung-cancer}. MTL brings a new opportunity for such techniques: by leveraging relationships between tasks, MTL can use the available supervisory signals for one task to aid the learning of other tasks.

\textbf{Applications.}
MTL is currently being employed in Computer Vision (CV) due to its success in achieving advanced scene understanding. Its most studied area is urban scene detection \cite{multinet, end-to-end-autonomous-driving, unified-autonomous-driving, Andrej-talk}, specifically to address autonomous driving related tasks such as road segmentation and object detection. MTL has also been successfully used in robotics, specifically in robotic-assisted surgery \cite{AP-MTL,ST-MTL} to predict diverse effects from a surgery scene (instruments, tissues etc.). Additionally, this paradigm has been heavily studied in the context of face recognition \cite{face_recognition-facial-expression, face-parsing, privacy-preserving-face, MTLFace} to enable, for instance, the simultaneous prediction of facial expression, face detection, and identification. MTL has also been explored in medical applications, such as in medical image segmentation in the area of gastroenterology for detection of polyps \cite{upper-gastro, explicable-capsule-endoscopy, esophageal}, or in cardiology for atrial segmentation \cite{3d-cardiac}. In addition, MTL has been applied to non-CV scenarios such as Natural Language Processing (NLP) \cite{MTL-NLP-networks, NLP-MTL-ADAPTIVE, NLP-NER, Vietnamese-POS, NLP-SEQ, NLP-POS} and recommendation systems \cite{MTL-recommendation-systems-1, MTL-recommendation-systems-2, MTL-recommendation-systems-3}.

\begin{figure*}[t!]
\centering
\includegraphics[width=0.95\textwidth]{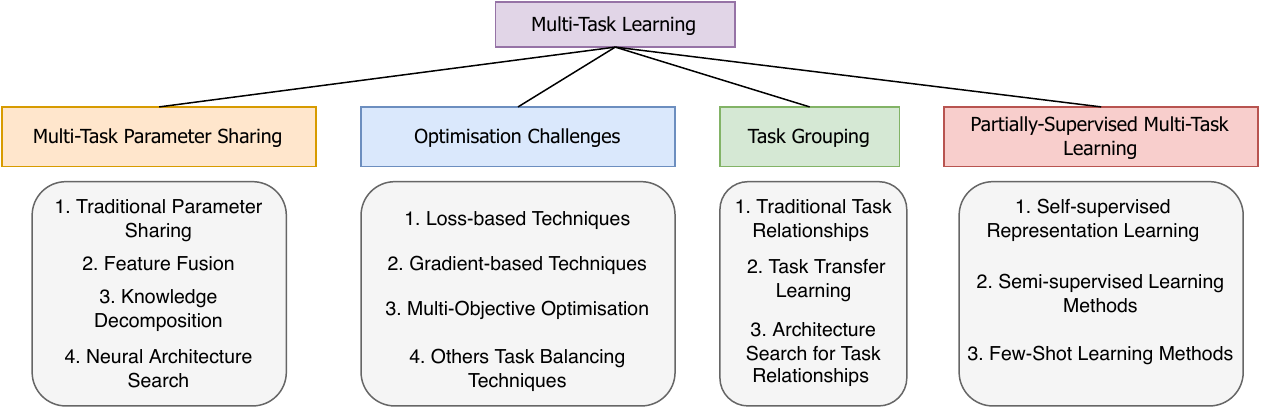}
\caption{Overview of the literature review structure. Firstly, we introduce Multi-Task Parameter Sharing in \cref{chapter:MT-parameter-sharing}. Secondly, we review Optimisation Challenges in \cref{chapter:Optimisation}. Thirdly, we review how task relationships can be be used to group them in \cref{sec:task-grouping}. Finally, we introduce, in \cref{chapter:partial-supervision}, the different partially-supervised computer vision methods in MTL.}
\label{overview}
\end{figure*}

\begin{figure}[t!]
\centering
\includegraphics[width=8.5cm]{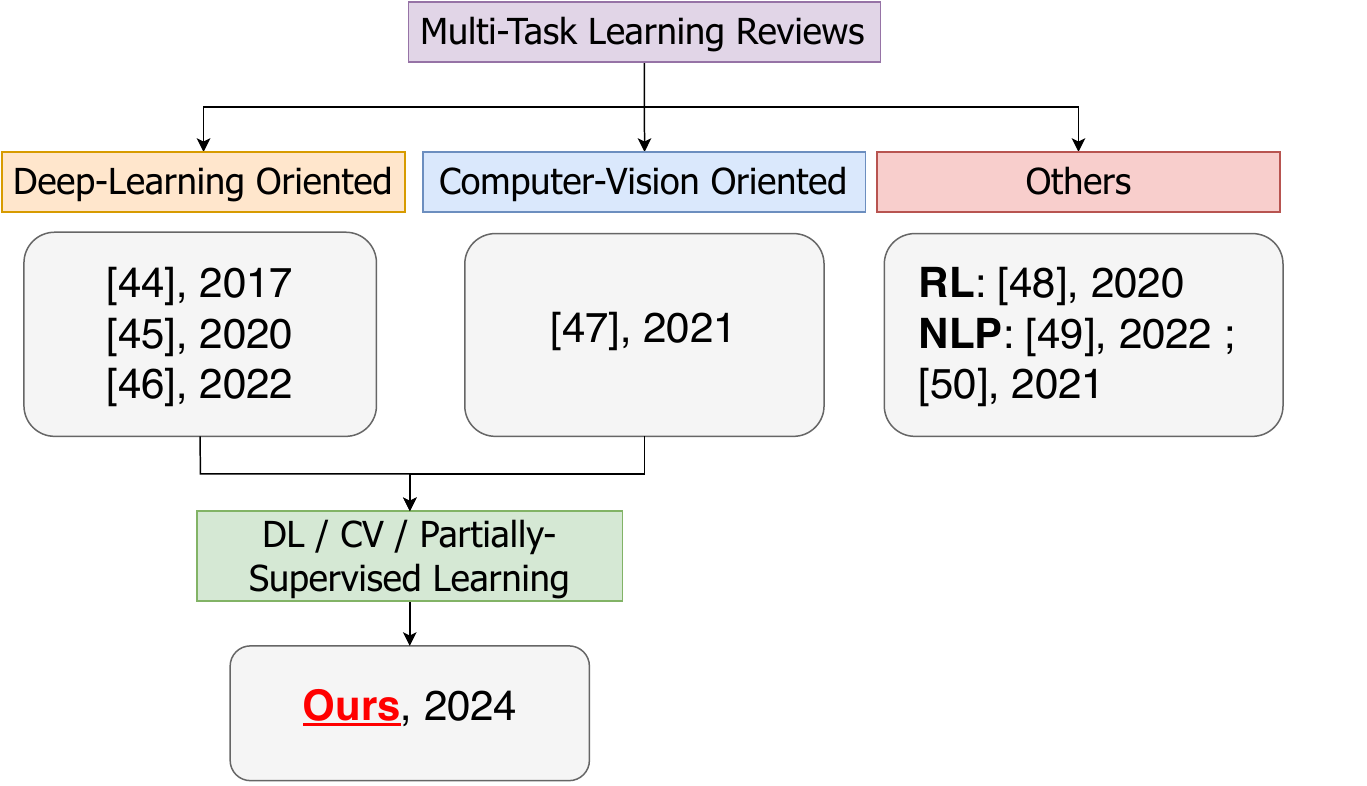}
\caption{Overview of the different reviews on Multi-Task Learning.}\label{related-work}
\end{figure}

\textbf{Related Work.} 
MTL has been the subject of numerous and diverse review papers \cite{MTL-overview-2017,MTL-in-DNN-2020, MTL-survey-2021,MTL-dense-predic-tasks, Deep-reinforcement-learning-survey, MTL-NLP-survey, MTL-NLP-overview}. Some of these previous works have focused on specific domains such as NLP. For instance, \citet{MTL-NLP-overview} focus on MTL-based solutions for various NLP tasks and provide a classification for available solutions, whilst \citet{MTL-NLP-survey} focus on NLP-related training procedures and task relatedness. Alternatively, \citet{Deep-reinforcement-learning-survey} review MTL in the domain of deep reinforcement learning (DRL). Other published reviews have focused on MTL from an optimisation perspective, for instance by comparing the different loss weighting techniques \cite{a-comparison-of-loss-weightin-strategies} or by evaluating task-specific transfer learning strategies \cite{taskonomy, factors-of-influence}. 

Some works have, however, aimed at providing a less constrained review of MTL. For instance, \cite{MTL-overview-2017, MTL-in-DNN-2020} reviews fully-supervised MTL methods as well as the inherent optimisation challenges under the deep learning framework. Moreover, \cite{MTL-survey-2021} provides a full-fledged and comprehensive review on both linear and DL solutions as well as the underlying optimisation techniques.

Previous work has focused on area more closely related to this study. For example, a CV-focused review \cite{MTL-dense-predic-tasks} analyses how MTL has been applied to pixel-wise prediction tasks and provides benchmark results on common fully-supervised MTL architecture. \cite{MTL-dense-predic-tasks} further differentiates MTL architectures based on the location where task interactions take place (encoder vs decoder). This paper, in contrast, does not highlight such differentiation as architectural issues are not the focus of our analysis. This review instead focuses on partially-supervised learning paradigms applied to CV tasks in a multi-task fashion. Although the vast majority of multi-task learning solutions has been applied to dense prediction tasks, this work aims at providing a comprehensive understanding of how MTL's future improvement might be underpinned by increasing the number and diversity of tasks. This study is the first, to the best of our knowledge, to focus on partially-supervised MTL for CV. 
\\
The literature for this review was selected through a comprehensive search of academic databases and was further refined based on the relevance to the central themes of this study and the author's expert judgment, ensuring the inclusion of both foundational and cutting-edge research of significant interest.

\textbf{Paper overview.} \cref{chapter:MT-parameter-sharing} reviews traditional fully-supervised MTL methods from a parameter-sharing perspective. 
\cref{chapter:Optimisation} introduces challenges arising from such multi-objective optimisation. 
\cref{sec:task-grouping} analyses relationships between common CV tasks, and how task groupings can be used to identify mutually beneficial tasks. 
\cref{chapter:partial-supervision} discusses how MTL can be used under partially-supervised paradigms. Last, \cref{sec:datasets} is dedicated to an introduction to available datasets, code repositories and tools as well as a comparison of the solutions introduced in this review.
We provide an structural overview of this work in \cref{overview}.

Furthermore, we provide an outline of the varied landscape of related MTL reviews, contextualizing our research within this framework. See \cref{related-work} for more details.

\section{Multi-Task Parameter Sharing}
\label{chapter:MT-parameter-sharing}
In order to understand the underlying challenges to MTL, \cref{sec:non-neural-MTL} reviews cross-task parameter sharing introduced in traditional settings. Subsequently, \cref{sec:feature-fusion} will review feature fusion paradigms under two major frameworks: CNN and Vision Transformers. Then, \cref{sec:knowledge-decomposition} investigates how learned representations can be partitioned and further shared. Finally, \cref{sec:parameter-NAS} will focus on architecture search based strategies as a way to share parameters across different tasks. 

\subsection{Traditional Parameter Sharing}
\label{sec:non-neural-MTL}
\subsubsection{Sparse Multi-Task Representations}
\label{sec:sparse}
The core of the early work in MTL has focused on obtaining a sparse multi-task parameter matrix generally obtained by linear models such as support vector machines (SVM) or ridge regression. Concretely, a parameter matrix is said to be sparse if a large proportion of its values are close to 0. The sparsity objective is based on the assumption that only a low-dimensional sub-representation of parameters should be shared across all the tasks. For example, \textit{Multi-Task Feature Learning} (MTFL) \cite{MT-feature-learning} defines the objective as an optimisation using the L1 regularisation. Considering a linear feature matrix $U \in \mathbb{R}^{d \times d}$ where $d$ is the parameter dimension, MTFL \cite{MT-feature-learning} aims at learning a transformation matrix $A \in \mathbb{R}^{d \times T}$ where $T$ is the number of tasks, such that $W$ = $UA$, with $W \in \mathbb{R}^{d \times T}$. Formally, such objective can be defined as the minimisation of the following function:
\begin{equation} \label{eq:sparse}
f(A, U) = \sum_{t=1}^{T} \sum_{i=1}^{m} L(y_{ti}, a_{t} \cdot (U^{T}x_{ti})) + \gamma||A||^{2}_{1},
\end{equation}
where the first term is the empirical error for the $i^{th}$ data-label pair $(x_{ti}, y_{ti})$ for a task $t$. In the second term, the transformation matrix $A$ is constrained by the the regularisation term, which is itself controlled by the non-negative parameter $\gamma$. As a result, the sparsity imposed on the transformation matrix $A$ will lead to most rows in $A$ being equal to 0. After the transformation $W = UA$, these rows will represent task-specific parameters whilst others represent the shared low-dimensional subspace $W$ across tasks. 
However, such objective only partition features. MTFL \cite{MT-feature-learning} aims to jointly learn the parameters and their partition. The resulting strategy is therefore to minimise the function $f$ over the parameter $U$. However, although such strategy results in a bi-convex on $A$ and $U$ individually, the minimisation optimisation objective is not, rendering the optimisation challenging. Therefore, MTFL \cite{MT-feature-learning} introduces a convex formulation to their problem. To a further extent, the authors suggest non-linear features can be obtained through the use of kernel learning \cite{learning-multiple-tasks-with-kernel-methods} therefore allowing the model to learn non-linear relationship between parameters. 

Following this sparsity objective, previous work has investigated using different linear models such as the Group Lasso Method \cite{taking-advantage-of-sparsity}, by improving over the convergence speed of the sparsity objective, or by minimising the trace-norm of $A$ \cite{MTL-l21, trace-norm-minimization}. 
Nonetheless, this paradigm is essentially constrained to only a small subset of shared features. Moreover, it also assumes tasks are related as some features are shared anyway. However, intuition suggests it should not always be the case. To counter this, some works \cite{dirty-model, learning-task-grouping-and-overlap} allow for an adaptive and partial overlapping of the task parameters to only share parameters when necessary.

\subsubsection{Clustering}
\label{sec:clustering}
To mitigate the a-priori assumption that all tasks are related, some works have investigated how to identify task relationships under a task clustering framework. Such methods are motivated by the assumption that similar tasks have similar weight vectors. Obtaining such clusters helps narrow down the search space for the shared low-dimensional parameter space.
For instance, \citet{clustering-learning-tasks} introduce a Task Clustering (TC) algorithm based on K-Nearest Neighbours (KNN) in which information is shared within clusters. Specifically, given two tasks $T_{1}$ and $T_{2}$, performance gain (PG) is calculated for the task pair through transfer learning (\ie $PG_{T_{1} \rightarrow T_{2}}$ if knowledge is transferred from $T_{1}$ to $T_{2}$). The task clusters are formed based on such pair-wise performance gains. Then, knowledge transfer is performed only within the most related tasks.
Similarly, \citet{dirichlet-process-priors} introduce an automatic identification of such clusters based on the Dirichlet Process (DP) prior distribution. 
Later, with the aim of providing a convex formulation to this framework, \citet{clustered-MTL} suggest regularising the multi-task parameter space $W$ by imposing 3 different norms to model several orthogonal properties: the mean weight vector size $\Omega_{mean}$ which measures how large the weight vectors are on average by computing the trace over the $T$-task weight representation, 
\begin{equation} \label{eq:clustering1}
\Omega_{mean}(W) = tr(WUW^{T}),
\end{equation}
where $U \in \mathbb{R}^{T \times T}$ is a projection matrix which has all its entries equal $\frac{1}{T}$. Subsequently, the between-cluster and the within-cluster variance which respectively measures how close together the clusters are and how dense the clusters are. These measures can be formulated as follows:
\begin{equation} \label{eq:clustering2}
\Omega_{between}(W) = tr(W(M-U) W^{T}),
\end{equation}
\begin{equation} \label{eq:clustering3}
\Omega_{within}(W) = tr(W(I-M) W^{T}),
\end{equation}
where $M = L-I$ for which $L$ is the laplacian matrix and $I$ is an identity matrix. 
Finally, \citet{clustered-MTL} choose to combine these measures through a weighted sum as part of their minimisation objective : 
\begin{equation} \label{eq:clustering}
\min 
\left\{
\sum_{y \in {\{mean, between, within}\}} \gamma_{y}\Omega_{y}(W)
\right\},
\end{equation}
where $\lambda$ is a weight parameter for the norm $\Omega$ over the weight matrix $W$.
This multi-criteria weighting leads to a decomposition of $W$ such that similar tasks are close in parameter space. 

To explicitly model the distributions of the tasks to better identify their relationships, \citet{kernels-for-MTL,learning-a-kernel} introduce a kernel learning strategy to find a Reproducing Kernel Hilbert Space (RKHS) in which task-respective distributions are close together in parameter space if their relatedness is high enough. 
Finally, \citet{CMTL-via-ASO} interestingly derive the relationships between Clustered Multi-Task Learning (CMTL) in which similar tasks are clustered and sparse multi-task representations are learnt within clusters, as seen in \cref{sec:sparse}. The work introduces three algorithms to perform CMTL and demonstrates how the clustering approach is significantly more efficient than the low-dimensional subspace learning solution, especially under high-dimensional data settings.

\subsubsection{Common-Trunk}
\begin{figure}[t!]
\centering
\includegraphics[width=8.5cm]{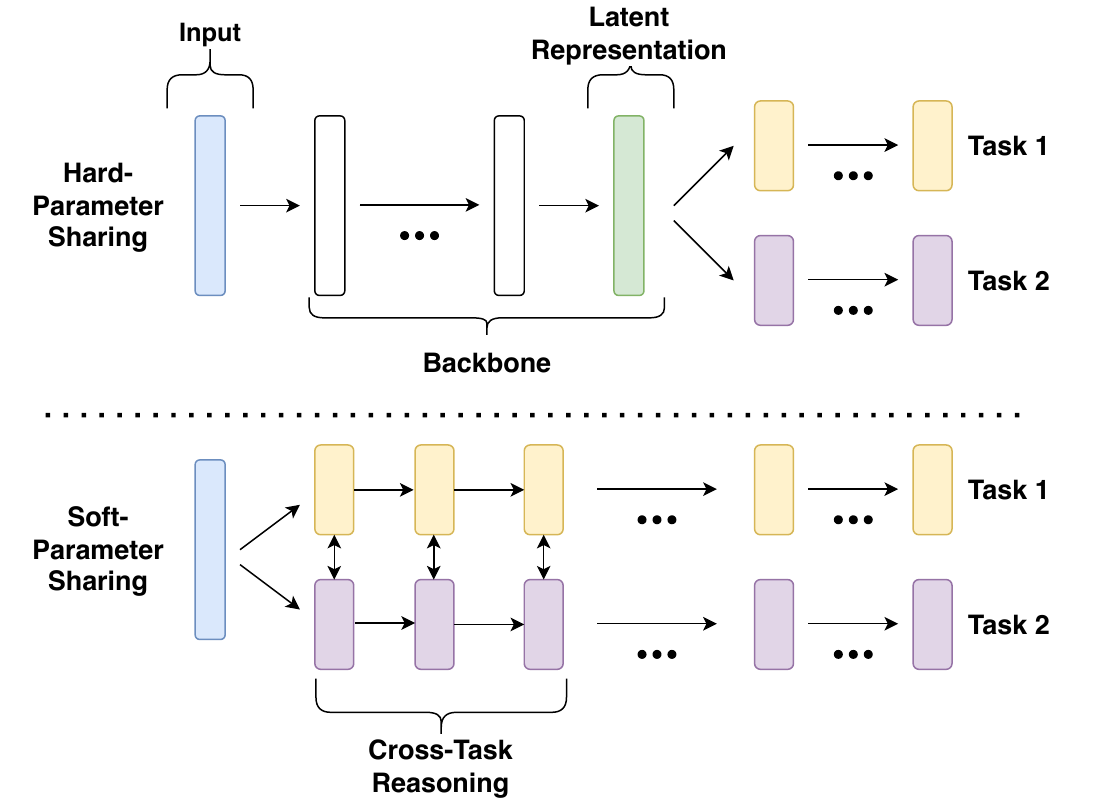}
\caption{Multi-Task Learning has mainly been divided into two architectural design schemes. Hard-parameter sharing (top) splits a shared backbone into task-specific heads which receives input from from the same set of features. Soft-parameter sharing (bottom) uses task-specific networks, but allows information to be shared between them.}\label{soft-hard}
\end{figure}
Early DL methods involved attaching task-specific heads to a CNN encoder's latent representation as shown in \cref{soft-hard} (top) \cite{Caruana}. For example, Ubernet \cite{Ubernet} introduced a CNN designed to tackle seven CV tasks. Many subsequent studies followed this design \cite{3d-cardiac, uncertainty, multinet, multitask-centernet}. This architecture shares a CNN backbone which gets updated by gradients aggregated by multiple tasks. As a result, all the tasks pull features from this backbone, which makes a global learned representation critical, although not trivial to obtain as different tasks need different representations to perform well. 
Hence, recent works suggest sharing parameters as part of multi-task encoder-decoder architectures at the decoder level \cite{PAD-net, PAP, MTI-NET, exploring-relational-context, invPT} to exchange high-level semantic features. For instance, Prediction-and-distillation Network (PAD-NET) \cite{PAD-net} suggests sharing knowledge after predictions and allows the training of a distillation module to learn what to share. \citet{MTI-NET} expend on this idea whilst incorporating multi-scale prediction for better dense prediction task performance. Similarly, at the prediction level, Pattern-Affinitive-Propagation (PAP) \cite{PAP} proposes learning pair-wise task relationship to produce affinity matrices for each task to further guide the sharing strategy.  

\subsection{Feature Fusion}
\label{sec:feature-fusion}
This section introduces parameter fusion techniques used in the two most pre-dominant vision models. First, \cref{subsec:CNN} introduces methods to share parameters across CNNs. Then, \cref{subsec:transformer} reviews recent attention-based methods to fuse parameters in Vision Transformers (ViTs) \cite{ViT}.
\subsubsection{CNN Sharing Strategies}
\label{subsec:CNN}
Cross-stitch Networks \cite{cross-stich} introduce a model-agnostic fusion technique. As opposed to the hard-parameter sharing paradigm, in which task-decoders are attached to a shared backbone encoder (Fig.~\ref{soft-hard} (top)), \citet{cross-stich} introduce a soft-parameter sharing paradigm in which task networks are processed independently and through which parameter fusion is executed in parallel at a similar level of abstraction (Fig.~\ref{soft-hard} (bottom)). Given two task activation maps $A$ and $B$, cross-stitch units \cite{cross-stich} compute the dot product between a vector representing their respective values $x^{i,j}_{A}$ and $x^{i,j}_{B}$  at a shared location (i, j) and a trainable weight matrix $W \in \mathbb{R}^{k \times k}$ , where \textit{k} is the number of tasks. The values in $W$ represent task-specific (diagonal entries) and shared parameters (non-diagonal entries). The process for $k=2$ is illustrated as:

\begin{equation} \label{eq:cross-stitch}
\begin{bmatrix}
\tilde{x}^{i,j}_{A}\\
\tilde{x}^{i,j}_{B}
\end{bmatrix}
= 
\begin{bmatrix}
w_{AA} & w_{AB}\\
w_{BA} & w_{BB}
\end{bmatrix}
\begin{bmatrix}
x^{i,j}_{A}\\
x^{i,j}_{B}
\end{bmatrix}.
\end{equation}
Despite being a locally-flexible, easy-to-implement and model-agnostic method, its design results in a complex and expensive model. First, studies investigating CNN layers have shown that shallow layers are usually task-agnostic and cross-stitch units would eventually represent such task-agnostic parameters, but at an expensive of training cost.  Second, the overall solution is expensive as the training costs increase with the number of tasks and the size of the network.

\begin{figure}[t]
\centering
\includegraphics[width=8.5cm]{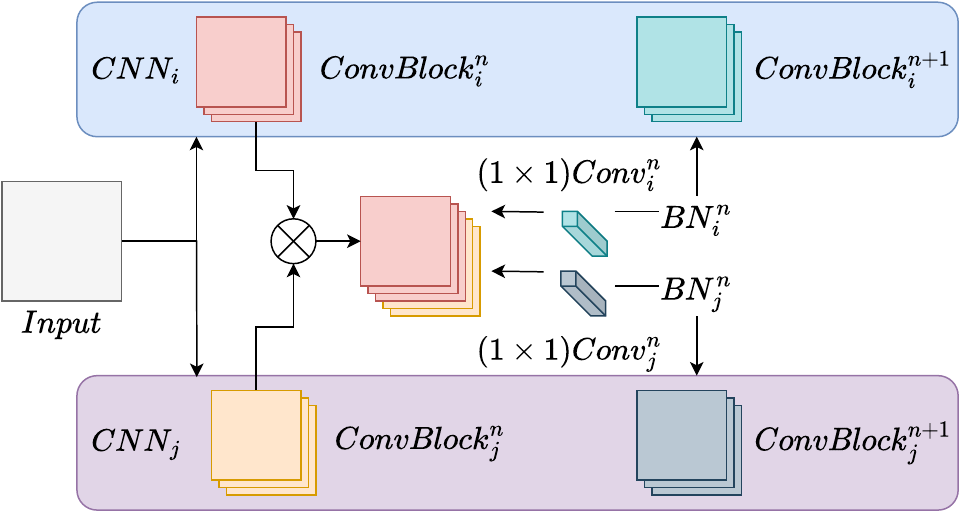}
\caption{Two task-specific CNN models CNN$_{i}$ and CNN$_{j}$. The NDDR-layer \cite{NDDR} first concatenates the representations of the respective convolutional blocks.
$1 \times 1$ convolutions are then run on this concatenation, one per task. Last, after batch normalisation, the features are propagated on to the next convolutional block of each model.
}\label{NDDR}
\end{figure}

Sluice Networks \cite{sluice} generalise cross-stitch units by increasing the flexibility and sharing parameter space. In particular, \citet{sluice} allow for selective sharing of layers, parameter subspaces and skip connections.
To expand on this soft-parameter sharing structure, \citet{NDDR} propose a solution based on the principle of Neural Discriminative Dimensionality Reduction (NDDR). This principle attempts to minimise the number of features whilst keeping the maximum amount of representative information, similarly to Linear Discriminant Analysis (LDA) or Principal Component Analysis (PCA). Therefore, NDDR \cite{NDDR} formulates the multi-task feature fusion problem as a discriminative dimensionality reduction problem by first concatenating parallel feature maps, then task-specific $1 \times 1$ convolutions \cite{network-in-network} are run on such representation to perform dimensionality reduction. In addition, the authors employ batch normalisation and weight decay to stabilise learning. This method is illustrated in \cref{NDDR}.

As a result, NDDR \cite{NDDR} enables learnable local representation parameter sharing in a similar manner to cross-stitch and sluice networks \cite{cross-stich, sluice}. However, these techniques hypothesise that all tasks should be processed together, computational cost could therefore be reduced using prior knowledge on task groupings to avoid redundant computation. 


\subsubsection{Attention-based Sharing Strategies}
\label{subsec:transformer}
With the advent of the transformer model \cite{attention-is-all-you-need}, originally applied to NLP, and subsequently to CV \cite{ViT}, there has been a great improvement in dense prediction tasks in CV \cite{PVT, PVT-v2, swin-transformer, focal-transformer} due to the non-local feature acquisition inherent to these models as well as their capacity to exploit long-range dependencies. 
Similar to the aforementioned soft-parameter sharing techniques \cite{cross-stich, sluice, NDDR}, Multi-Task Attention Network (MTAN) first trains a single CNN network which is designed to learn general features. Then, task-specific networks are derived by attaching attention modules, which learn soft-attention masks over the shared features, to each convolutional operation of the aforementioned CNN network.

\textit{Unified Transformer} (UniT) \cite{UniT} learns a multi-modal encoder-decoder transformer model. UniT \cite{UniT} learns modality-specific encoders using multi-head self-attention, the modalities are then simply concatenated before a joint decoder performs cross-attention to mix the multiple representations. Similarly, \textit{Multi-Task Transformer Network} \cite{MulT} (MulT) performs feature fusion at the decoding level and introduces a shared attention mechanism. Specifically, MulT chooses a reference task $t^{ref}$,
then the reference task encoded representation $x$ is used to compute a query $q^{t^{ref}}_{x}$ and a key $k^{t^{ref}}_{x}$. Let us denote $v^{t}$ the values for the other tasks based on the previous stage output. The attention values for this task are then calculated as:
\begin{equation} \label{eq:attention}
A^{t^{ref}}_{x} = softmax\left(\frac{q^{t^{ref}}_{x} \cdot {k^{t^{ref}}_{x}}^{T}}{\sqrt{C_{qkv}^{t^{ref}}}}\right) + B^{t^{ref}}.\
\end{equation}
Subsequently, for any task $t$, the shared representation is obtained as:  $\tilde{x}^{t} = A^{t^{ref}}_{x}v^{t}$.  The term $x^{t}$ is then used for the multi-head attention.

MTFormer \cite{MTFormer} also chooses to compute cross-task interactions at task-specific heads. However, the authors choose to concatenate the projected representations at each transformer block, based on multi-head self-attention operations. To merge the attention maps of $n$ tasks, the authors show it is beneficial to consider self-task attention as a primary task and to consider cross-task attention as playing an auxiliary role in order to perform cross-task feature propagation. To reflect this, the authors choose to reduce the number of projected feature channels $C$ of auxiliary tasks such that $C' = \frac{C}{n-1}$, whilst keeping the original dimension for the main task. 

Finally, motivated by the success of pyramid-based transformer-based encoded representations for dense prediction tasks \cite{swin-transformer, PVT, PVT-v2}, InvPT \cite{invPT} proposes a cross-scale self-attention mechanism for multiple tasks. In this method, the attention maps are linearly combined by learnable weights, the result is also constrained by a residual feature map from the input image.

\subsection{Knowledge Decomposition}
\label{sec:knowledge-decomposition}
Knowledge Decomposition aims at partitioning a large set of features into smaller and meaningful components. In the context of MTL, one might be interested in recycling large models into smaller multi-task models. 
First, \cref{sec:tensor-factorization} reviews how tensor factorization can operate over CNN kernels to construct MTL components. Second, \cref{sec:knowledge-distillation} introduces methods to transfer information from a large single-task teacher model to a smaller multi-task student model. 
Last, \cref{sec:adapters} reviews how adapters can be used to achieve multi-task continual learning by fine-tuning a large single-task model.  

\subsubsection{Tensor Factorization}
\label{sec:tensor-factorization}
\cref{sec:sparse} reviewed solutions employing the low-rank approximation of a multi-task weight matrix using linear models. Deep Multi-Task Representation Learning (DMTRL) \cite{deep-multi-task-representation-learning} generalises this idea to tensors (N-dimension arrays with $N \in \mathbb{N}$ and more specifically $N \ge 3$). In fact, as per the nature of a CNN, kernels are N-dimension tensors and fully convolutional (FC) layers are 2-way tensors, stacking those by a number of tasks $T$, usually resulting in large tensors. Tensor Factorization (TF) is a generalisation of some form of matrix decomposition, such as Singular Value Decomposition (SVD) \cite{SVD}.
DMTRL \cite{deep-multi-task-representation-learning} accomplishes soft-parameter sharing in a layer-wise manner between parallel and identical CNNs, similarly to \cite{cross-stich, NDDR}. First, single-task CNNs are trained, then layer-wise parameters are concatenated during backpropagation and subsequently fed as input to SVD-based solutions for decomposition. DMTRL \cite{deep-multi-task-representation-learning} uses multiple sharing strategies, including one based on the Tucker Decomposition (TD) \cite{Tuck1966c}, to learn parameters of this SVD-based solution to generate the decomposed units. 

Further to this strategy, \citet{Trace-norm-DML} use the tensor trace norm (the sum of a tensor's singular values) as a proxy of the tensor rank on the layer-wise parameters' concatenation. In this way, each CNN is encouraged to use the other network's parameters. However, these methods have the same drawback as the previously introduced parameter-fusion based techniques \cite{cross-stich, NDDR, sluice} as parameters are shared in a layer-wise fashion which introduces constraints including architectural parallelism and locality in the parameter sharing strategy. 

\subsubsection{Knowledge Distillation}
\label{sec:knowledge-distillation}
Another perspective to parameter sharing is to design strategies based on Knowledge Distillation (KD). KD is a form a model compression that transfers knowledge from a large model to a smaller model. Early KD work in MTL explored how to compress DRL methods. For instance, \cite{policy-distillation, actor-mimic, distral} introduced a policy distillation strategy to derive lighter multi-task policies from task-specific deep Q-network (DQN) policies. However, as per the nature of DRL, these strategies approach tasks for which the set of actions was finite and would therefore struggle in more complex prediction visual tasks. 
As a result, \citet{PAD-net} introduce, as part of a multi-task multi-modal network, a distillation module to merge predictions from intermediate and complementary tasks from different modalities to subsequently pass representations on to task-specific decoders. The variations for this distillation module include cross-prediction reasoning as well as attention-guided mechanisms. 
Hence, \citet{knowledge-distillation-for-mtl} suggest a two-step solution in which: (1) task-specific models are first trained before freezing their respective parameters; (2) a multi-task model is optimized to minimise a multi-loss objective through the use of \textit{adaptors} (reviewed in \cref{sec:adapters}) that align task-specific and task-agnostic parameters together in order for the multi-task model to use the same features as the task-specific models. Following a similar strategy, \citet{MuST} extend this strategy to a self-supervised pre-training procedure through the use of intermediate pseudo-labeling. 

Recently, \citet{factorizing-knowledge-in-NNs} introduce a new alternative to KD, namely, \textit{Knowledge Factorization} (KF). Instead of distilling knowledge from a task-specific teacher model to a multi-task student model, KF aims at decomposing a pre-trained, large multi-task model into \textit{k} task-disentangled factor networks modelling both task-agnostic and task-specific parameters of the teacher model. The resulting lightweight networks can be assembled to create custom multi-task models.

\subsubsection{Adapters}
\label{sec:adapters}
With the aim of learning universal representations that can perform well across multiple domains, \citet{learning-multiple-visual-domains} introduce \textit{residual adapter modules}. Adapters are small neural networks that learn to recognise task-specific parameters given a model pre-trained on another task. Inspired by the ResNet \cite{resnet} architecture where residual connections are introduced across the sequential process of a CNN, adapters are modules attached after each convolutional block that learn to select parameters to be utilised for a downstream task. This presents an alternative to traditional \textit{fine-tuning} as only the adapters are trained. \citet{learning-multiple-visual-domains} demonstrate the capacity of adapter modules to maintain performance across 10 domains by just tuning a small portion of domain-specific parameters, and also their capacity to overcome the challenge of \textit{learning without forgetting} \cite{learning-without-forgetting}. 

\citet{efficient-parametrization-of-multi-domain-nn} introduce \textit{parallel} adapters as a simpler variant and show that only a few parameters need to be re-trained. As opposed to domain learning, \citet{attentive-single-tasking} show how adapters can be used in Incremental Multi-Task Learning (I-MTL). As a new task is optimised, \citet{attentive-single-tasking} train task-specific adapters to identify what parameters to retrain and \textit{Squeeze-and-Excitation} \cite{squeeze-and-excitation} modulation blocks perform channel-wise attention. Furthermore, to address the challenges raised by I-MTL, AdapterFusion \cite{adapter-fusion}, inspired by the multi-task objective adapter training strategy proposed by \cite{bert-and-pals}, introduces a 2-stage algorithm that enables task-specific parameters inside a transformer model to re-use other task-specific parameters contained in adapters. 
It is worth noting that, apart from the few aforementioned studies, adaptors have been studied far less in CV than in NLP. There is thus scope for exploiting this efficient parameter-sharing more fully in CV applications


\subsection{Neural Architecture Search}
\label{sec:parameter-NAS}
Neural Architecture Search (NAS) generally attempts to find the best network architecture given a specific problem by manipulating neural modules. However, in case of a multi-task objective, NAS can be seen as a way to partition the parameter space. For instance, \citet{soft-layer-ordering} introduce parameter sharing through \textit{soft ordering} (as opposed to \textit{parallel ordering}). The idea is to learn individual weight scalars per shared layers to \textit{soft-merge} parameters at different depths of a network. This comes down to learning a N-dimension tensor of task-specific parameters. Alternatively, Multi-gate Mixture of Experts (MMoE) \cite{MMoE} embeds the Mixture of Experts (MoE) framework \cite{MoE-layer} in MTL by sharing expert task-specific networks and optimising a gating network to select what features to use for each task. Following the same framework, \citet{DSelect-k} further improve the efficiency and stability of the selection of experts process and demonstrates its significant improvement on large-scale multi-task datasets. 

With the aim of learning an even more flexible assembling strategy, evolutionary algorithms have been proposed as a training strategy in which agents are network inference routes consisting of a set of computational blocks \cite{path-net, evolutionary-architecture-search-CTR}. Similarly, to learn large-scale MTL systems that tackle \textit{catastrophic forgetting} in the I-MTL paradigm, \citet{dynamic-introduction-of-tasks-in-large-scale} adopt an evolutionary algorithm to dynamically optimise a model each time a new task is added. Moreover, motivated by even more flexible ways to share features, some work has investigated using computational operations inherent to CNN layers as modulation units. For instance, \citet{flexible-mtl-by-learning-parameter-allocation} introduce the Gumbel-SoftMax Matrix model by modulating inner components of a layer, and shows how their activation is learned to optimise tasks through logits. Alternatively, \citet{adashare} show how routing policies can be learned through the Gumbel-Softmax sampling method \cite{gumbel-softmax} taking into account computational resources. Recently, \citet{AutoMTL} use this trick to integrate the learning of such policies into its programming framework. \citet{stochastic-filter-groups} modulate networks the same way as \cite{gumbel-softmax}, however \textit{stochastic filter groups} are introduced as a  way to model the distributions, approximated via \textit{variational inference} \cite{variational-inference}, over the possible kernel groupings. More recently, Adashare \cite{adashare} introduced MTL in such architecture search-based systems to model relationships between tasks by studying the partitioned feature space. As a result, recent studies have focused on leveraging these relationships to route information through networks. For instance, \citet{fully-adaptive-feature-sharing} incrementally expand on an initially small network, at each step, grouping similar tasks based on a measure of task affinity. Similarly, \citet{branched-mtl} implement a low-resource, layer-wise sharing strategy driven by NAS, exploiting task affinity measures. In a CV context, \citet{synergistic-MTL-NAS} leverage hardware-aware NAS \cite{h-NAS-survey} together with MTL to improve the accuracy of dense-prediction tasks on edge devices.
\\
\\
\section{Optimisation Challenges}
\label{chapter:Optimisation}
MTL has underlying optimisation challenges due to it being a Multi-Objective problem. MTL is subject to two major optimization issues. First, overall performance is dependent on the relatedness of the tasks being optimised. Unrelated tasks can have conflicting gradients that will lead to a non-convergence of a multi-task solution. This phenomenon is called \textit{negative transfer}. 
Second, multi-objective performance relies on a thorough task balancing problem as respective complexities interfere during training. For example, easier tasks converge faster, resulting in larger gradients being back-propagated across all tasks. This makes the acquisition of aggregated representations for different task gradients non-trivial.
This section reviews solutions aiming to tackle the aforementioned challenges. 
First, \cref{sec:loss-based-methods} reviews how individual losses can be adjusted to balance tasks in a MTL training. Second, \cref{sec:grad-based-methods} focuses on techniques that directly operate over gradient updates during training. Third, \cref{sec:moo} reviews techniques directly inspired from Multi-Objective Optimisation to perform gradient-descent under a MTL configuration. Last, \cref{sec:other-task-balancing-techs} introduces other task balancing approaches.

\begin{figure*}[t!]
\centering
\includegraphics[width=0.8\textwidth]{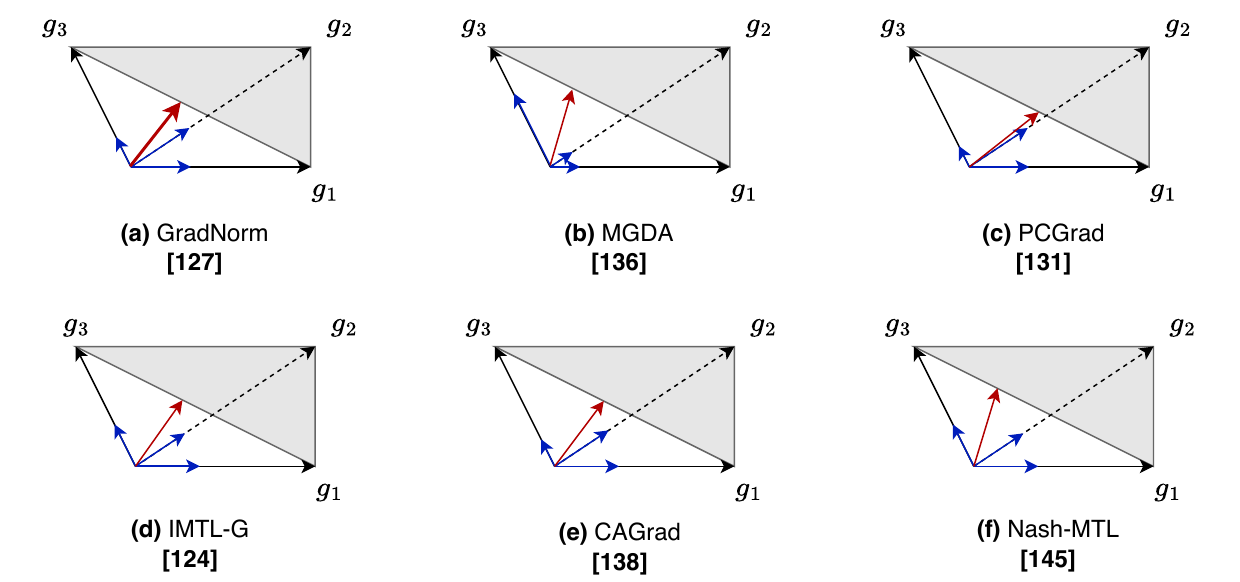}
\caption{Visualisation of the different gradient update methods in MTL. The \textcolor{blue}{blue} arrows represent the projections of the task-specific gradient update noted as $g_{1}$, $g_{2}$ and $g_{3}$. The \textcolor{red}{red} arrow represents the aggregated gradient update. \label{optimisation-techs}}
\end{figure*}

\subsection{Loss-based Techniques}
\label{sec:loss-based-methods}
Early work in deep MTL studied a weighted average of the task-specific losses $L_{i}$. By considering $\mathit{T}$ tasks, this multi-task loss can be formulated as follows:
\begin{equation} \label{eq:basic-mtl}
L_{MTL} = \sum_{i=1}^{T} w_{i} L_{i},
\end{equation}
where $\mathit{w_{i}}$ are respective positive task weights.

Rather than setting weights manually, solutions have been proposed to incorporate the weights into the objective function to adaptively weigh tasks during training. For example, AdaLoss \cite{adaloss} suggests adaptively tweaking weights in such a way that they are inversely proportional to the average of each loss in order to project losses onto the same scale. Alternatively, \cite{uncertainty} introduces learnable scalar parameters into the minimisation objective. The authors derive their loss weighting strategy based on the \textit{Homoscedastic (or task-dependent) uncertainty} which captures the uncertainty of a model, this type of uncertainty is invariant to different inputs. The authors follow a Gaussian likelihood maximisation setting and show that the loss optimisation given two tasks can be approximated as:
\begin{equation} \label{eq:uncertainty}
L_{uncert} (W, \sigma_{1}, \sigma_{2}) = \frac{1}{2\sigma_{1}^{2}} L_{1} + \frac{1}{2\sigma_{2}^{2}} L_{2} + \log \sigma_{1} \sigma_{2}.
\end{equation}
Following the same strategy, \citet{auxiliary-tasks-in-MTL} suggest a slight difference in the $\log$ regularisation term, by changing it to $\log(1+\sigma^{2})$. This is to prevent values of $\sigma \in [0,1]$ yielding negative loss values. We refer to this method as \textit{revised uncertainty}.
However, uncertainty-based task balancing strategies have certain drawbacks and in practice, task-wise terms need to be changed in \cref{eq:uncertainty} depending on the type of task (classification or regression) and depending on the task-specific loss.  As a result, IMTL \cite{impartial-MTL} introduces a hybrid method using both gradient methods and adaptive loss tuning. The loss component IMTL-L updates task-specific parameters and learns task-wise scaling parameters $s$ by minimising a function $g$ for each task as:
\begin{equation} \label{eq:IMTL-L}
g(s) = e^{s}L(\theta) - s.
\end{equation}
\cref{eq:IMTL-L} shows that each task loss is scaled by $e^{s}$ and regularised by $s$ to avoid trivial solutions. In practice, this technique allows tasks to all have comparable scales. Moreover, as opposed to uncertainty weighting \cite{uncertainty}, IMTL-L does not bias towards any type of task such as regression or classification. 
Alternatively, \citet{geometric} introduce a Geometric Loss Strategy (GLS), using the geometric loss to weigh $n$ task-specific losses $L_{1...n}$. The geometric loss is invariant to individual loss scales which makes it an easy way to balance tasks. As a result, \citet{geometric} decide to weight respective tasks as follows:
\begin{equation} \label{eq:geometric}
L_{geometric} = \Pi_{i=1}^{n} \sqrt[n]{L_{i}}.
\end{equation}
Additionally, the authors introduce a variant to focus on $m$ ($m < n$) `more important' tasks and therefore attribute more weighting to these as demonstrated below:
\begin{equation} \label{eq:geometric-prime}
\Tilde{L}_{geometric} = \Pi_{j=1}^{m} \sqrt[m]{L'_{j}} \times \Pi_{i=1}^{n} \sqrt[n]{L''_{i}}.
\end{equation}

Alternatively, balance of tasks can be achieved through averaging task weights over time by considering the rate of change in the respective task-specific loss. \citet{MTAN} introduce \textit{Dynamic Weight Average} (DWA). DWA calculates a specific task-specific weight $\lambda_{k}$ for a task $k$ by obtaining a relative descending rate compared to other tasks with respect to the previous iteration (averaged over multiple epochs) as follows:
\begin{equation} \label{eq:DWA}
\lambda_{k}(t) = \frac{K exp(w_{k}(t-1) / T)}{\sum_{i}exp(w_{i}(t-1) / T)} , w_{k}(t-1) = \frac{L_{k}(t-1)}{L_{k}(t-2)},
\end{equation}
where $T$ is a temperature parameter controlling the stiffness of the weighting distribution and $K$ ensures that $\sum_{i}\lambda_{i}(t) = K$.

More recently, Random Loss Weighting (RLW) \cite{RLW-RLG} has drawn task-specific weights from a probability distribution at each epoch before normalising them and shows comparable results to state-of-the-art (SOTA) loss-weighting strategies. As a result, \citet{RLW-RLG} provide a more generalisable solution than the baseline (\cref{eq:basic-mtl}), due to its additional randomness. Finally, \cite{a-comparison-of-loss-weightin-strategies} provides benchmark results comparing Single Task Learning (STL) to DWA \cite{MTAN}, uncertainty \cite{uncertainty} and revised uncertainty \cite{auxiliary-tasks-in-MTL} and suggests that, given careful task selection, the revised uncertainty method \cite{auxiliary-tasks-in-MTL} generally performs best but suffers when there is lack of training samples.

\subsection{Gradient-based Techniques}
\label{sec:grad-based-methods}
Weighting losses is an indirect way of changing the model's gradients. Therefore, a line of work has investigated how to optimise MTL models by directly operating over the gradients. Throughout this section, we refer to the illustration in \cref{optimisation-techs} which provides a visualisation of the gradient update techniques introduced by the presented methods. Informally, the problem is that during multi-task optimisation, a subset of parameters $\theta$ is shared across multiple tasks, as a result, $\theta$ generally receives gradient updates to optimise all tasks at once. In practice, this is achieved by finding an aggregated representation of the vectors. However, finding such representation is not trivial as task-respective gradients might conflict. Hence, GradNorm \cite{gradnorm} proposes a method that balances training by automatically tuning the gradient magnitudes. Considering a subspace of weights of a model $W$ (generally chosen as the last shared layer for computational purposes), GradNorm \cite{gradnorm} defines the $L_{2}$ norm of the gradient for a particular weighted task loss $\textit{i}$, and similarly defines $\overline{G_{W}}(t)$ the average gradient norm across all tasks at time $t$. Additionally, GradNorm \cite{gradnorm} defines 2 training rates. The first training rate is task-specific and is defined as $\widetilde{L}_{i}(t) = L_{i}(t)/L(0)$. It is the loss ratio for a task $i$ at time $t$. The second training rate defines a relative training rate for a task $i$ as follows:
\begin{equation} \label{eq:training-rate}
r_{i}(t) = \widetilde{L}_{i}(t) / \sum_{i=1}^{n} \widetilde{L}_{i}(t),
\end{equation}
where the right term is an averaged training rate over all tasks $n$ for the given time $t$.

Subsequently, GradNorm \cite{gradnorm} calculates new task-specific gradients for the weight subspace $W$ based on the update rule below:
\begin{equation} \label{eq:gradnorm}
G_{W}^{i}(t) = \overline{G}_{W}(t) \times [r_{i}(t)]^{\alpha}, 
\end{equation}
where $\alpha$ is a hyper-parameter controlling the force of traction towards a similar training rate for all tasks. This method, by directly operating over gradients during training, adaptively tunes the speed to which tasks are being trained. 
However, solely balancing tasks does not prevent conflicting gradients (negative transfer).

\begin{figure}[t!]
\centering
\includegraphics[width=8cm]{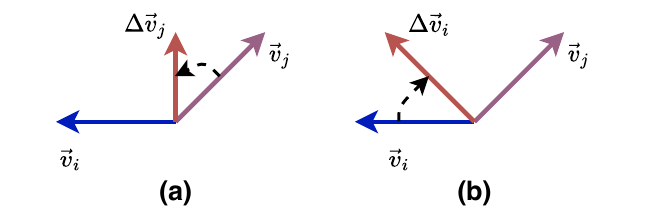}
\caption{Conflicting vectors (with negative cosine similarity), where $\vec{v}_{i}$ and $\vec{v}_{j}$ represent task-specific updates. In case (\textbf{a}) PCgrad \cite{gradient-surgery} projects $\vec{v}_{j}$ onto the normal space of $\vec{v}_{i}$ resulting in $\Delta\vec{v}_{j}$. In case (\textbf{b}), PCgrad \cite{gradient-surgery} oppositely projects $\vec{v}_{i}$ onto the normal space of $\vec{v}_{j}$. The resulting projections are later added to update the model parameters.
\label{PCgrad-figure}}
\end{figure}

GradDrop \cite{GradDrop} proposes adding a modular layer that operates during back-propagation to first select a sign (positive or negative) based on the initial distribution of gradient values. It then proposes masking out all gradient values of the opposite sign. 
Similarly, \citet{adapting-aux-losses-using-gradient-similarity} leverage auxiliary tasks in order to optimise a main task. During training, \citet{adapting-aux-losses-using-gradient-similarity} only minimise the auxiliary losses if their gradient update at epoch $t$ is non-conflicting with the main task gradient update. Specifically, \citet{adapting-aux-losses-using-gradient-similarity} use the \textit{cosine similarity} to measure the gradients relation. Conceptually, if the cosine similarity between the main and auxiliary gradients is positive, it suggests that the auxiliary loss should be minimised alongside the main loss, otherwise, it should not. \citet{regularising-DMTL-orthogonal-gradients} use a similar strategy in a more conventional MTL setting, in which multiple tasks are optimised simultaneously. \citet{regularising-DMTL-orthogonal-gradients} use the cosine similarity to ensure shared gradients are near orthogonal. The authors refer to conflicting gradients when these have a negative cosine similarity, and non-conflicting when it is positive. Unlike \cite{regularising-DMTL-orthogonal-gradients} which ensures `near orthogonal' properties of the gradients via the minimisation of the loss, PCGrad \cite{gradient-surgery} projects only conflicting gradients by projecting those of task $i$ onto the normal plane of task $j$ as shown in \cref{PCgrad-figure} (b). Formally, such projection can be defined as:
\begin{equation} \label{eq:pcgrad}
\Delta g_{i} = g_{i} - \frac{g_{i} \cdot g_{j}}{\lVert g_{j} \rVert^{2}} g_{j}.
\end{equation}
However, imposing such strong orthogonality constraint upon gradients implies that all tasks at hand should benefit from similar gradient interactions, ignoring complex relationships and destructing natural optimisation behaviour. Moreover, PCGrad \cite{gradient-surgery} stays idle when the gradients have positive cosine similarity, which still might not be optimal as a more desirable similarity (closer a positive cosine similarity) might be preferred. Hence, GradVac \cite{gradient-vaccine} leverages both directions and magnitudes in an adaptive strategy. Specifically, given two tasks $i$ and $j$, a similarity goal $\phi^{T}_{i,j}$ is fixed between two gradients $\mathbf{g_{i}}$ and $\mathbf{g_{j}}$ such that $\phi^{T}_{i,j} > \phi_{i,j}$ for which $\phi_{i,j}$ is the cosine similarity, as computed in PCGrad \cite{gradient-surgery}. To achieve this, GradVac \cite{gradient-vaccine} derives the projection equation (\cref{eq:pcgrad} by fixing the gradient of $\mathbf{g_{i}}$ and rather estimates the weight of $\mathbf{g_{j}}$ via the Law of Sines in the gradients plane. This process can be summarised as:

\begin{equation} \label{eq:vcgrad}
\Delta g_{i} = g_{i} + \frac{ \lVert g_{i} \rVert (\phi_{ij} \sqrt{1 - \phi_{ij}^{2}} - \phi_{ij} \sqrt{1 - (\phi_{ij}^{T})^{2})}}{\lVert g_{j} \rVert \sqrt{1 - (\phi_{ij}^{T})^{2}}} \cdot g_{j}.
\end{equation}

Furthermore, using an Exponential Moving Average (EMA) (similar to DWA \cite{MTAN}), $\phi^{T}_{ij}$ is estimated in an adaptive manner during training, over a subset of shared parameters $\mathbf{w}$ belonging to the same layer as:

\begin{equation} \label{eq:ema-update-vcgrad}
\Delta \phi_{ijw} = (1 - \beta) \phi^{t}_{ijw} + \beta \phi^{t-1}_{ijw}.
\end{equation}

Similarly, \citet{impartial-MTL} suggest a hybrid method leveraging both loss and gradient tweaking, IMTL \cite{impartial-MTL} chooses, in their gradient component IMTL-G, to make all the projections from each task equal to balance the tasks. 
Recently, RotoGrad \cite{rotograd} proposed a solution to both homogenise gradients magnitude and resolve conflicting ones. To achieve this, a 2-step algorithm is implemented. The first step consists in homogenising the gradients such that the tasks that have progressed the least are encouraged to learn more. Therefore, to project the gradients $\mathbf{G_{k}}$, for a task $k$, Rotograd \cite{rotograd} assigns weights to gradients such that their weighted combination is $\mathbf{C = \sum_{k} \alpha_{k} \lVert G_{k} \rVert}$. Precisely, $\alpha$ is adaptively calculated every $i^{th}$ iteration as:

\begin{equation} \label{eq:rotograd-alpha-update}
\alpha_{k} = \frac{\lVert G_{k} \rVert / \lVert G_{k}^{0} \rVert}{\sum_{i} \lVert G_{i} \rVert / \lVert G_{i}^{0} \rVert}.
\end{equation}

In the second step, Rotograd \cite{rotograd} tunes the gradients by learning a task-specific rotation matrix $\mathbf{R_{k}}$ on the last shared representation $\mathbf{z}$. Hence, $\mathbf{R_{k}}$ aims to maximise the cosine similarity between the gradients across tasks given a batch of size $n$; or equivalently, to minimise the loss function. This process can be illustrated as:

\begin{equation} \label{eq:rotograd}
L_{rot}^{k} = - \sum \left\langle R^{T}_{k} g_{n,k} , v_{k} \right\rangle.
\end{equation}

\subsection{Multi-Objective Optimisation}
\label{sec:moo}
Multi-Objective Optimisation (MOO) addresses the challenge of optimising a set of possibly conflicting objectives. This section reviews gradient-based multi-objective optimisation methods applied to MTL. First, \cref{sec:pareto-optimality} formally defines Pareto optimisation and how it is relevant to MTL under gradient descent techniques. Then, \cref{sec:gradient-descent-solutions} reviews gradient-descent optimisation solutions applied to MTL.

\subsubsection{Pareto Optimality}
\label{sec:pareto-optimality}
As presented in \cref{sec:loss-based-methods}, tuning the task-specific weights is not trivial and usually comes at the cost of computational overhead. One way to remedy this is to reframe the MTL optimisation into a MOO problem.
Motivation to use MOO for MTL comes from the fact that global optimality for multiple tasks is unconceivable unless a pair-wise equivalence between tasks exists, which is unrealistic. For a hard-parameter sharing network as depicted in \cref{soft-hard} (top), $\theta^{sh}$ represents parameters that are shared across all tasks and $\theta^{t}$, $t \in T$, are task-specific parameters. Additionally, $\widehat{L^{t}}(\theta^{sh}, \theta^{t})$ is the empirical loss for a specific task $t \in T$. Then, a multi-objective loss function can be defined as:

\begin{equation} \label{eq:moo-minimisation}
\min_{\theta^{1}, ... ,\theta^{T}} (\widehat{L}^{1}(\theta^{sh},\theta^{1}), ... , \widehat{L}^{T}(\theta^{sh},\theta^{T})).
\end{equation}

\begin{figure*}[t!]
\centering
\includegraphics[width=0.8\textwidth]{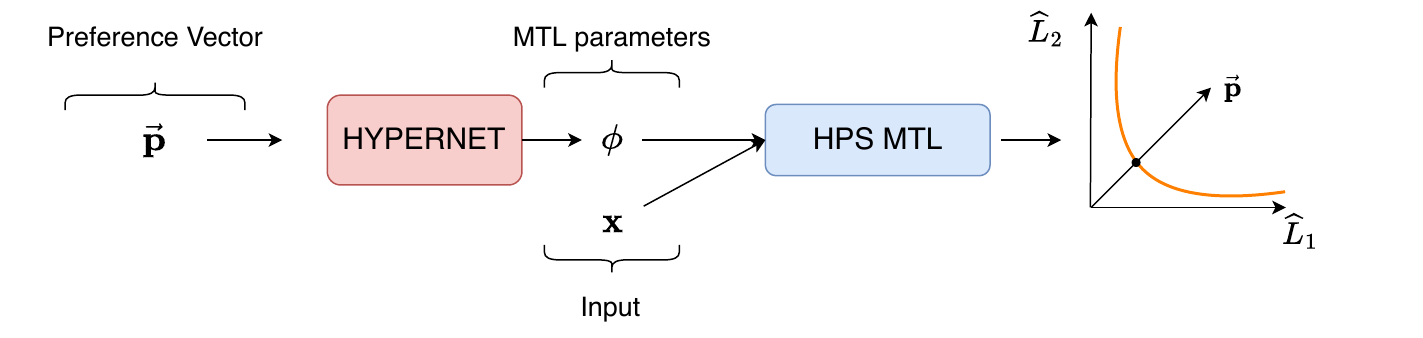}
\caption{Controllable Pareto MTL \cite{controllable-pareto-mtl} approximates the Pareto and allows for real-time trade-off optimisation. A preference vector is sampled using Monte-Carlo methods. This vector is given as input to a Hypernet which outputs the parameters of Hard-Parameter Sharing (HPS) model. The returned Pareto optimal solution is the closest to the preference vector.  \label{controllable-pareto-fig}}
\end{figure*}

\noindent Minimising \cref{eq:moo-minimisation} leads to Pareto-optimal solutions. In other words, in a MTL setting, considering both shared and task-specific parameters $\theta_{i}^{sh,t}$ and $\theta_{j}^{sh,t}$ for task $i$ and $j$ respectively, a Pareto-optimal solution is one for which a change in $\theta_{i}^{sh,t}$ would damage the performance of task $j$ and vice-versa. The set of Pareto-optimal solutions can therefore be considered as a set of trade-offs between tasks \cite{navon2021learning}. This set is called the \textit{Pareto front} ($P_{\theta}$).

Pareto optimality has extensively been studied leveraging the Multiple Gradient Descent Algorithm (MGDA) \cite{MGDA_paper} which supports the Karush-Khun-Tucker (KKT) conditions that are necessary conditions for Pareto optimality. 
MGDA \cite{MGDA_paper} demonstrates that minimising \cref{eq:minimisation-mgda} supports the KKT constraints and states that the result of this minimisation is either 0 and therefore results in a multi-task solution which satisfies the KKT conditions (a point along the pareto front); otherwise, this minimisation leads to a descent direction that improves all tasks. This process can be depicted as:
\begin{equation} \label{eq:minimisation-mgda}
\min_{\alpha^{1}, ...,\alpha^{T}} 
\left\{
\mynorm{ \sum_{t=1}^{T}\alpha^{t}\nabla_{\theta^{sh}}\widehat{L}^{t}(\theta^{sh}, \theta^{t})}^{2}_{2} 
\right\},
\end{equation}
where $\alpha^{t}$ are non-negative scaling factors such that: $\sum^{T}_{t} \alpha^{t} = 1$.

\subsubsection{Gradient Descent Solutions}
\label{sec:gradient-descent-solutions}
In a MTL context, \citet{MTL-as-MOO} show that MTL optimisation can be regarded as a MOO problem using MGDA and demonstrates that solving \cref{eq:moo-minimisation} is equivalent to finding the min-norm point in the convex hull formed by the input points. That is, finding the closest point in a convex hull to a query point. As a result, \citet{MTL-as-MOO} obtain the aggregated projection of the task-specific gradient vector updates. Subsequently, to solve \cref{eq:moo-minimisation}, \citet{MTL-as-MOO} use the Frank-Wolfe solver \cite{Frank-Wolfe} and ensures, with negligible additional training time, the convergence to a Pareto-optimal solution.    
CAGrad \cite{conflict-averse-gradient-descent-for-mtl} generalises the MGDA algorithm and chooses to ensure the convergence of the MTL objective to the equally weighted average of task-respective losses. To achieve this, CAGrad \cite{conflict-averse-gradient-descent-for-mtl} first obtains an average vector $d$ of individual task updates $g_{i}$. Then, it aims to find an update vector $g_{w}^{t}$ on a pre-defined ball around $d$, which maximises the worst local improvement between ${T}$ tasks defined as: $\max_{d\in \mathbb{R}} \min_{i \in {T}} \langle g_{i},d \rangle$. This way, CAGrad \cite{conflict-averse-gradient-descent-for-mtl} balances the different task-specific objectives. Furthermore, the authors show the dominance of CAGrad in a semi-supervised setting compared to MGDA \cite{MTL-as-MOO}.
However, this approach ensures the convergence to any point along the Pareto front which might not be representative of the desired task balance, an unbalanced solution might be preferred to enhance a target task.
Therefore, Pareto MTL \cite{Pareto-MTL} proposes generalising MGDA to generate a set of multiple Pareto optimal solutions along the Pareto front which would serve as different trade-offs to choose from. To achieve this, \citet{Pareto-MTL} take inspiration from \cite{mmo-decomposition} and decomposes the objective space into $K$ well-distributed unit preference vectors $u_{k}$ to guide solutions. Formally, this is achieved through a sub-problem to \cref{eq:moo-minimisation} where a dot-product maximisation constraint is imposed between $u_{k}$ and a given vector $v$ to guide the learning onto a targeted area of the Pareto front. A sub-region is defined as:
\begin{equation} \label{eq:partition-pareto}
\Omega_{k} = \{v \in \mathbb{R}_{+}^{m} | u_{k}^{T}v \leq u_{k}^{T}v, \forall j = 1, ..., K\}.
\end{equation}

In contrast to its predecessors, \citet{continuous-pareto-exploration-in-MTL} suggest generating continuous Pareto optimal solutions along the Pareto front. To achieve this, \citet{continuous-pareto-exploration-in-MTL} propose a 2-stage training algorithm that, in its first stage, generates a single Pareto stationary point $x_{0}$ from a network's initialisation. Then, a set of points $x_{n}$ is explored along the tangent plane direction $v_{i}$ and the points are calculated as: $x_{i} = x_{0} + sv_{i}$ where $s$ is a step size. As a result, a set of directions is obtained. Finally, \citet{continuous-pareto-exploration-in-MTL} combine the tangent vectors acquired in the previous step through linear combination to form convex hulls in which Pareto solutions are obtained, resulting in a continuous approximation of a larger Pareto front. 

All the solutions introduced thus far in this section initialise network parameters per trade-off, resulting in a large storage demand and making solutions computationally inefficient. Additionally, the generated solutions are either singular \cite{MTL-as-MOO} or subject to the practitioner's preferences \cite{Pareto-MTL, continuous-pareto-exploration-in-MTL}. To alleviate both issues, \citet{controllable-pareto-mtl} propose utilising a HyperNet \cite{HyperNetworks}, a type of neural network that learns to generate the weights of another network, rendering storage less demanding. Additionally, \citet{controllable-pareto-mtl} introduce preference-based training to perform trade-off selection in real-time. More specifically, the objective space is sampled into $K$ subspaces (similarly to \cite{continuous-pareto-exploration-in-MTL}). Specifically, given a preference vector $\mathbf{p}$, the goal is to find a local Pareto optimal solution within such subspace for which the angle is the smallest to $\mathbf{p}$. To train the network on representative trade-off preference vectors, vectors are sampled using Monte Carlo methods and are given as input to the HyperNetwork $\mathbf{G}$. \citet{controllable-pareto-mtl} use a standard hard-parameter sharing strategy and such a process is depicted in \cref{controllable-pareto-fig}. Similarly, along the lines of preference-driven Pareto optimal solution, \citet{multi-objective-multi-task-learning-framework-induced-by-pareto} choose to directly cast the MOO optimisation as a Weighted Chebyshev (WC) problem which consists of finding the Pareto front by minimising the $l_{+\infty}$-norm between the initialisation point and the Pareto front. 

Recently, Nash-MTL \cite{bargaining-game} suggests a different approach to obtain an Pareto optimal solution. Inspired by the game theory literature, the authors directly aim at obtaining the \textit{Nash Bargaining Solution} \cite{Nash-bargaining-solution} which can be found on the Pareto front and translates to a proportionally fair solution where any change to the state results in a negative update for at least one task. Specifically, let's consider $U \in \mathbb{R}^{T}$ the set of all possible trade-offs and similarly, $ D \in \mathbb{R}^{T}$ the default set of disagreements, namely, a trade-off if all tasks $T$ fail to agree on an agreement. Moreover, in order to find a task agreement, namely, find a solution for $U$ with columns $u_{i}$ such that $\forall_{i} : u_{i} > d_{i}$, the authors demonstrate that finding a Nash Bargaining Solution is equivalent to solving:
\begin{equation} \label{eq:nash-solution}
\begin{aligned}
u^{*} = \arg \max_{u \in U} \sum_{i=1}^{T} \log (u_{i} - d_{i}) \\
\textrm{s.t.} \forall_{i} : u_{i} > d_{i}
\end{aligned}
\end{equation}
Subsequently, the authors propose an iterative solution to solve \cref{eq:nash-solution} and find the aggregated update vector is equivalent to solving \cref{alg:algo}

\begin{algorithm}
\DontPrintSemicolon
\caption{Nash-MTL}\label{alg:algo}
\KwIn{$\theta^{0}$, an initial parameter vector; $\eta$, learning rate}
    \For{\texttt{t= 1,...,T}}{ 
     Computer task-specific gradients: $g^{t}$ \;
     Let $G^{t}$ be a matrix with columns $g^{t}$ \;
     Solve for $\alpha$: $(G^{t})^{\mathbf{T}}G^{t}\alpha = 1/\alpha$, to obtain $\alpha^{t}$ \;
     Update parameters: $\theta^{t} = \theta^{t} - \eta G^{t} \alpha^{t} $ \;
    }
\Return{$\theta^{T}$} \;
\end{algorithm}

where $G \in \mathbb{R}^{m \times T}$ is a multi-task gradient matrix with parameter dimension $m$. Moreover, $\alpha \in \mathbb{R}^{T}_{+}$ is a strictly positive matrix which acts as a constraint to the objective which conceptually renders gradient vectors in $G$ orthogonal when they need to be. Additionally, $t$ is a task iterator, $\theta$ represents the shared parameter networks, $\eta$ the learning rate. $G^{t}$ is a task-specific vector update matrix with columns $g^{t}_{i}, i \in T$. 
The results obtained by \cite{bargaining-game} suggest Nash-MTL achieves current state of the art weighting strategy under many MTL configurations.

However, recently, \citet{even-help} instead demonstrated that most MTL optimisation strategies \cite{MTL-as-MOO, gradnorm, gradient-surgery, GradDrop} do not improve MTL training beyond what careful choice of scalar weights in MTL weighted average (\cref{eq:basic-mtl} can achieve. Rather, \citet{even-help} identify MTL optimisation is particularly sensitive to the choice of hyper-parameters.

\subsection{Other Task Balancing Techniques}
\label{sec:other-task-balancing-techs}
\subsubsection{Stopping Criterion Techniques} 
Previous techniques balanced tasks either by finding a combination of the task weights or through gradient manipulation to prevent destructive learning. However, these techniques globally penalise some tasks over others by constraining certain parameters in the objective space. 
Therefore, \citet{task-wise-early-stopping}, as part of their solution leveraging multiple auxiliary tasks to perform facial landmark detection, propose a task-wise early stopping strategy. The intuition is that once a task starts to overfit a dataset, it will harm the main task as it will force the optimisation to be stuck in a non-global optimum. Hence, a task is stopped if its performance, measured as the product between the training error tendency, noted as $L_{tr}$ and the generalisation error \textit{w.r.t} $L_{tr}$, noted as $L_{val}$, has not exceeded a certain threshold $\epsilon$. 
Formally, a training error rate $E_{tr}$ is calculated over a patience epoch length $k$ \textit{w.r.t} a current epoch $t$. Intuitively, the smaller $E_{tr}$, the greater the signal to continue the training for the task as the training loss substantially drops over the period of time $k$ as:
\begin{equation} \label{eq:early-stopping-tr}
E_{tr} = \frac{k \cdot med_{j=t-k}^{t}L_{tr}(j)}{\sum_{j=t-k}^{t}L_{tr}(j) - k \cdot med_{j=t-k}^{t}L_{tr}(j)},
\end{equation}
where $med$ represents the median operation. 
Similarly, $E_{val}$ measures the overfitting w.r.t $L_{tr}$. \citet{task-wise-early-stopping} define $\lambda$ as an additional learnable parameter to measure the importance of the task's loss. This process is shown in \cref{eq:early-stopping-val} below: \\
\begin{equation} \label{eq:early-stopping-val}
E_{val} = \frac{L_{val}(t) - \min_{j = 1..t}L_{tr}(j)}{\lambda \cdot \min_{j = 1..t}L_{tr}(j)}. 
\end{equation}
Overall, if $E_{tr} \cdot E_{val} > \epsilon$, the stopping criterion is met.

In a MTL configuration in which all the tasks are aimed to be optimised equally, stopping a task might result in \textit{catastrophic forgetting}. Therefore, \citet{12-in-1} propose a simple dynamic Stop-and-Go procedure that continually checks for task-wise improvement and degradation during training. 
Precisely, if performance, measured as the task-wise validation loss term for a given epoch $n$, noted as $L_{t}^{n}$, has not met the performance threshold $\epsilon_{stop}$ over the patience parameter $k$ such that $L_{t}^{n \rightarrow k} < \epsilon_{stop}$. Then, task $t$ is set to \textit{STOP} mode.
If during \textit{STOP} mode, $L_{t}^{n}$ is degraded and meets the degradation threshold $\epsilon_{go}$ such that $L_{t}^{n} < \epsilon_{go}$, then task $t$ is set back within the MTL training and is set to \textit{GO} mode.
In \cite{12-in-1}, the authors set $\epsilon_{stop}$ to be 0.1\% and $\epsilon_{go}$ to be a degradation of 0.5\% of the task's best performance.

\subsubsection{Prioritisation Techniques}
An alternative to balancing the learning of multiple tasks simultaneously is to instead focus on easier or complex tasks to benefit the training for all tasks. For example, \citet{self-paced-MTL} choose to guide their MTL training by gradually incorporating both harder tasks and harder instances into the objective function. By considering a number of tasks $T$ and a number of instances per task $n$, the authors propose a regularisation $f$ over $\mathbf{W} \in \mathbb{R}^{n \times T}$ as shown below:
\begin{equation} \label{eq:self-paced}
f(W, \lambda, \gamma) = - \lambda \sum_{i=1}^{T} \lVert W \rVert_{1} + \gamma \sum_{i=1}^{T} \frac{\lVert W \rVert_{2}}{\sqrt{n_{i}}},
\end{equation}
in which the first term imposes the negative $L_{1}$-norm on the instances $n$. This term prioritise easier instances over harder ones when $\lambda$ is low. This is motivated by the fact that easy instances, for which the empirical loss will be small, have bigger gradients. This behaviour is caused by the sparsity norm defined above. On the contrary, difficult  instances have bigger empirical losses and therefore smaller gradients. As a result, as training continues, gradually increasing $\lambda$ will introduce more difficult instances by increasing the difficult task gradients. Similarly, the second term imposes the $L_{2-1}$-norm on the task-specific data instances $n_{i}$. This is motivated by the fact that harder tasks exhibit larger empirical losses and gradually reducing $\gamma$ will introduce harder tasks. This enables the training to smoothly progress whilst avoiding both inter-instance and inter-task possible conflicts.

On the other hand, some works have focused on starting with harder tasks to benefit easier tasks. For instance, \citet{dynamic-task-prio} propose a loss weighting strategy leveraging the \textit{focal loss} \cite{focal-loss} as defined below:
\begin{equation} \label{eq:original-cross-entropy}
FL(\mathbf{p},\gamma) = -(1-\mathbf{p})^{\gamma}\log(\mathbf{p}).
\end{equation}
The focal loss, described in \cref{eq:original-cross-entropy} is primarily intended for classification, \citet{dynamic-task-prio} suggest using key performance metrics (KPIs) per task $t$ (i.e. accuracy, average precision etc...) to generalise the method. Specifically, they adjust these task-specific KPIs $\kappa_{t}$ in an EMA approach as shown below: 
\begin{equation} \label{eq:FL}
\bar{\kappa}_{t}^{(\tau)} = \alpha \kappa_{t}^{(\tau)} + (1 - \alpha) \kappa_{t}^{(\tau - 1)},
\end{equation}
where $\alpha$ is a discount factor and $\tau$ is the iteration. 
Subsequently, the authors swaps original focal loss probability \textbf{p} (described in \cref{eq:original-cross-entropy}) for their KIPs $\bar{\kappa}_{t}^{(\tau)}$. As a result, the authors define a task difficulty as a combination of the task-specific loss and its respective KPI-based focal loss as:
\begin{equation} \label{eq:DTP}
L_{DTP} = \sum_{t=1}^{T} FL(\bar{\kappa_{t}};\gamma_{t}) \widehat{L}_{t}.
\end{equation}

Alternatively, \citet{learning-to-MTL-active-sampling} propose prioritising harder tasks through \textit{active sampling} (\ie choosing what data to train a model with at a particular time $t$ during training). More specifically, the model keeps track of two performance estimations: $t_{i}$ and $c_{i}$ which are a target performance and current performance, respectively, for a task $i$. The task performance is measured as follows: $m_{i} = \frac{t_{i} - c_{i}}{t_{i}}$, where a higher value of $m_{i}$ indicates the model is currently bad at task $i$. Therefore, to encourage the model to prioritise harder tasks, a task-wise sampling strategy is modeled by a distribution $p_{i}$ at every $k$ decision steps which is calculated as follows:
\begin{equation} \label{eq:MTL-active-sampling}
p_{i} = \frac{\exp^{\frac{m_{i}}{\tau}}}{\sum_{c=1}^{k} \exp^{\frac{m_{c}}{\tau}}}.
\end{equation}
Subsequently, the probability distribution is used to sample the next tasks throughout training.

\section{Task Grouping}
\label{sec:task-grouping}

As explained in \cref{chapter:Optimisation}, the overall performance of a MTL model heavily depends on the set of tasks. The optimisation space could be simplified by only processing related tasks together. This chapter focuses on how Task Relationship Learning (TRL) can be achieved through Task Grouping (TG). 

\begin{figure}[t!]
\centering
\includegraphics[width=8.5cm]{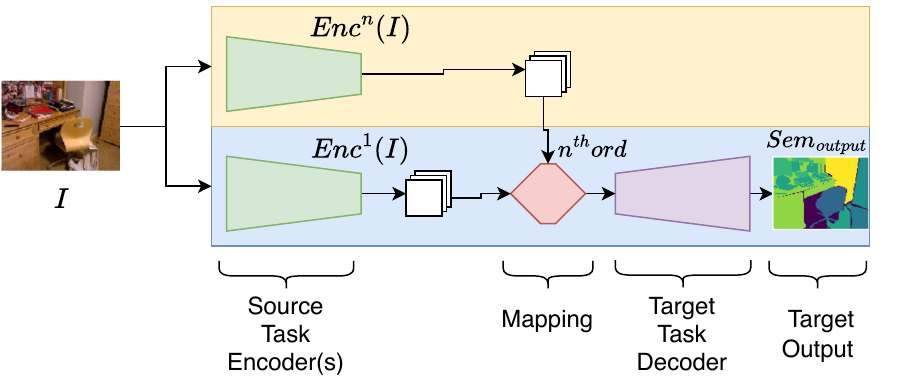}
\caption{Taskonomy \cite{taskonomy} leverages pre-trained encoder(s) (\textit{in green}) and estimates a mapping (\textit{in red}) from the latent representation to the input of the target task's decoder (\textit{in purple}).}
\label{transfer-mapping}
\end{figure}

Thus far, most works relied on human judgment concerning the relatedness of the tasks. However, these assumptions can be mitigated by quantitatively measuring task relationships. Early attempts in this area aimed to model task relationships (TR) based on vectors in a shared low-dimensional subspace. For example, \citet{learning-with-whom-to-share} explicitly build upon MTFL \cite{MT-feature-learning} (introduced in \cref{sec:non-neural-MTL}) and frames the task grouping problem as a mixed integer programming problem. GO-MTL \cite{learning-task-grouping-and-overlap} learns a linear combination of task-specific vectors. 
Later, \citet{MRN} expand on previous works modelling TRs using matrix-variate normal distribution over task-specific parameters regularisation techniques to identify positive task correlations \cite{MTRL}. However, to embed this regularisation technique into DL, \citet{MRN} use the tensor normal distribution \cite{tensor-normal-distribution} as a prior over task-specific tensors and learns task relationships by learning task covariance matrices. 
Similarly, \citet{MMoE} learn gating networks in a MoE framework to implicitly model task interactions. However, these works model relationships from a high-level perspective and generally poorly describe pair-wise relatedness.
To tackle TG, a body of work focused on studying relationship based on Transfer Task Learning (TTL) by directly learning a mapping between the learned parameters for a task $a$ to a target task $b$ in a MTL setting. For instance, Taskonomy \cite{taskonomy} introduced a computational approach to perform TG based on finding transfer learning dependencies between tasks. More specifically, \citet{taskonomy}, after training task-specific networks, the encoder parts of the networks are frozen and transfer task functions and dependencies are estimated via a target task decoder. Motivated by the idea that multiple source tasks can help provide a more meaningful dependency estimation for a mutual source task, the authors include high-order transfers where a mapping function receives the five best representation as inputs (from the five best first-order source tasks mappings), as illustrated in \cref{transfer-mapping}.
Additionally, Taskonomy \cite{taskonomy} derives a vision task clustering architecture and shows that 4 major clusters stand out, namely: 3D tasks, 2D tasks, low dimensional geometric tasks and semantic tasks. 
Calculating the affinities in such a way is extremely computationally expensive. To alleviate such demand for computation, as opposed to analysing the performance of TTL, Representation Similarity Analysis (RSA) \cite{RSA-taxonomy} directly investigates the feature maps learned by the task specific networks. The authors choose to leverage RSA to frame the task relationship problem by computing correlation through task-specific inferences on pairs of images. As a result, a dissimilarity matrix is obtained for each task-specific network and a similarity score is obtained through the Spearman’s correlation. 
However, these latter works only highlight the relationships from a transfer-learning perspective and do not present performance in a multi-task setting. 
Hence, \citet{which-tasks-should-be-learned-together} propose an alternative to transfer-learning based solutions to highlight task relationships. This alternative is motivated by two findings. First, results obtained by \citet{which-tasks-should-be-learned-together} do not show any correlation in the performances between measured \textit{task affinities} and multi-task learning setting. Second, transfer-learning affinities highlight high-level semantic dependencies as only the bottleneck of the source task encoder is used for the mapping. However, MTL should benefit from clean structural dependencies in all abstraction levels of the features. 
Instead, the authors frame this TG problem as an architecture search. Specifically, given an input image, the model aims to determine the best combinations of encoder backbones and task-specific decoders and perform an exhaustive search over these components. The process is constrained by a search time budget value given a number of tasks $T$. Moreover, \citet{which-tasks-should-be-learned-together} optimise the search space using a branch-and-bound procedure and trains between $\binom{T}{2} + T$ and $2^{T} - 1$ networks given $T$ tasks before performing TG. However, this search performance is computationally expensive and as a result, \citet{efficiently-identifying-task-groupings-for-MTL} directly build upon this framework and obtains task groups in a single run only. To achieve this, the authors introduce \textit{Task Affinity Grouping} (TAG) which is a \textit{look-ahead} algorithm that tracks changes in the MTL loss (in this case, \cref{eq:basic-mtl}) under different task groupings. Therefore, the authors introduce the notion of \textbf{task affinity} between two tasks $a$ and $b$ defined by $\mathcal{\hat{Z}}_{a \rightarrow b}$ as:
\begin{equation} \label{eq:affinity-a-b}
\mathcal{\hat{Z}}^{t}_{a \rightarrow b} = 1 - \frac{L_{b}(X^{t}, \theta^{t+1}_{s|a}, \theta^{t}_{b})}{L_{b}(X^{t}, \theta^{t}_{s}, \theta^{t}_{b})},
\end{equation}
in which $t$ is the step during the estimation procedure and where the loss $L_{b}$ for task $b$ is parameterised by $X, \theta_{s}, \theta_{b}$ which represents the input, the shared parameters and task-specific parameters for task b, respectively. The look-ahead term $\theta^{t+1}_{s|a}$ represents the update of the shared parameters w.r.t. the update on task $a$. Subsequently, a network selection procedure is implemented to maximise the total inter-task affinity score. For instance, for a set of tasks $\{T\}$ the affinity scores onto a task $a$ are averaged over all the tasks.
\begin{equation} \label{eq:overall_affinity}
\mathcal{Z}_{a} = \frac{\sum_{t}^{|T|}\mathcal{\hat{Z}}_{t \rightarrow a}}{|T|} , a \in \{T\}, t \neq a.
\end{equation}
This problem is NP-Hard and can therefore be solved by a branch-and-bound algorithm.

\section{Partially Supervised Multi-Task Learning}
\label{chapter:partial-supervision}
Methods reviewed so far have mainly focused on a fully-supervised setup which assumes that data is sufficient and all task labels are available. However, this setting is not always realistic as both acquiring data and task labels is an expensive process in certain cases. 
In practice, the diversity of the task set is limited as required data and labels generally do not co-exist within the same datasets and therefore, not in the same quantities and/or domains. Thus, there is a need to explore MTL in settings that utilise all available source of information. 
Thankfully, MTL systems can mitigate their data dependency by using available supervisory information of one task to enhance the training of the unlabelled tasks by leveraging task relationships. 
Therefore, in this chapter, \cref{sec:representation-learning} reviews how leveraging multiple auxiliary tasks in a self-supervised manner can help obtain a general representation tailored to downstream tasks. 
Then, \cref{sec:semi-supervised} studies MTL solutions in a semi-supervised settings in which all tasks are optimised.
Finally, \cref{sec:few-shot-learning} introduces how MTL can be framed in a low-data availability learning paradigm: Few-Shot Learning.
Throughout this chapter, we refer to \textit{Partial Supervision} as an umbrella term encompassing self-supervised learning, semi-supervised learning and few-shot learning.

\subsection{Self-Supervised Representation Learning}
\label{sec:representation-learning}
As seen in this review, finding a task-agnostic representation suitable for all the tasks is crucial. However, most previous work in MTL assumed high availability of data and focused on obtaining such representations without diminishing the demand for labels. To remedy this issue, an alternative way to obtain a shared representation is to exploit tasks in a self-supervised fashion. Self-supervised tasks are tasks for which labels can be created without manual annotations. Such tasks hold a strong advantage in the context of MTL as downstream tasks benefit from the representation induced by multiple tasks \cite{cross-stich}. As a result, Self-supervised Multi Task Learning (Self-MTL) can be leveraged as a pre-training strategy. 

For instance, \citet{MT-self-supervised-VL} suggest leveraging 4 self-supervised vision tasks as a pre-training procedure. \textit{Relative Position} \cite{relative-position} is a task which consists of finding the relative positions of a pair of patches sampled from the same unlabeled image. \citet{relative-position} claim to perform well at this task enhances object recognition. \textit{Colorization} \cite{colorization} which requires predicting the original RGB pixel color values given a greyscale image. This task acts as a cross-channel encoder and helps pixel-level dense prediction tasks. The `\textit{Exemplar}' task \cite{examplar-task} where pseudo-classes are estimated for each sample and the network is trained to discriminate between these. This task aims to improving classification properties in the learned representation. Last, \textit{Motion Segmentation} \cite{motion-segmentation} is a task that learns, given an image $I_{t}$ at a time $t$ to recognise pixels that will move in $I_{t+1}$. This task helps refine the features necessary to both object detection and segmentation prediction through movement cues. 

\begin{figure}[t!]
\centering
\includegraphics[width=8.5cm]{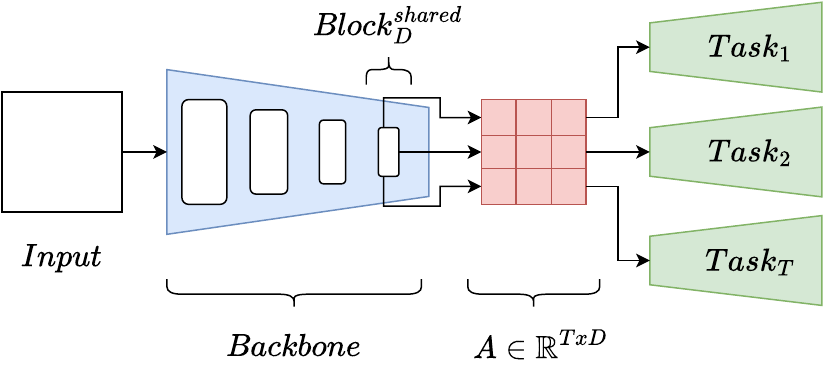}
\caption{In \cite{MT-self-supervised-VL}'s self-supervised solution, the task-specific heads (in green) receive a linear combination of high-level features from the last residual layers of a ResNet-101 \cite{resnet} encoder. The features are then selected via a matrix $\mathbf{A}$ which is trained to be sparse. This allows for a task-wise factorisation of the learned features to improve the generalisation of the CNN.}
\label{self-supervised-visual-learning}
\end{figure}

\citet{MT-self-supervised-VL} identify two possible sources of conflict in a Self-MTL setting. First, there are conflicts in the task respective inputs, as for instance, the colorization tasks receive greyscale images whilst others receive RGB images. This results in an network architectural problem. To resolve this conflict, the authors suggest performing \textit{input harmonisation} by duplicating the greyscale image over the RGB channels. Second, there is conflict in whether the features being trained should generalise to the class at hand or to the specific input image. To resolve this, the authors incorporate their CNN into a lasso regularisation block where each task-specific decoder receives a layer-wise linear combination of the shared backbone convolutional blocks. Hence, a matrix $A \in \mathbb{R}^{T \times D}$ is trained to be sparse where $T$ is the number of task-specific decoders and $D$ is the number of convolutional blocks being shared. This regularisation allows the network to factorise the features to enhance the generalisation of the network. The authors present results matching fully-supervised single-task performance on diverse CV tasks such as classification, detection and depth prediction. The authors' solution is illustrated in \cref{self-supervised-visual-learning}. 

MuST \cite{MuST} uses specialised teacher models to pseudo-label unlabeled multi-task datasets and suggests a pre-training strategy based on the following tasks: classification, detection, segmentation and depth estimation. Subsequently, a multi-task student model is trained on the pseudo-labeled dataset. Fine-tuning on downstream tasks shows that the self-supervised pre-training outperforms traditional ImageNet pre-training baseline \cite{ImageNet} and additionally, the authors identify that a large number of tasks and datasets benefit the representation for downstream tasks.

\begin{figure*}[t!]
\centering
\includegraphics[width=0.95\textwidth]{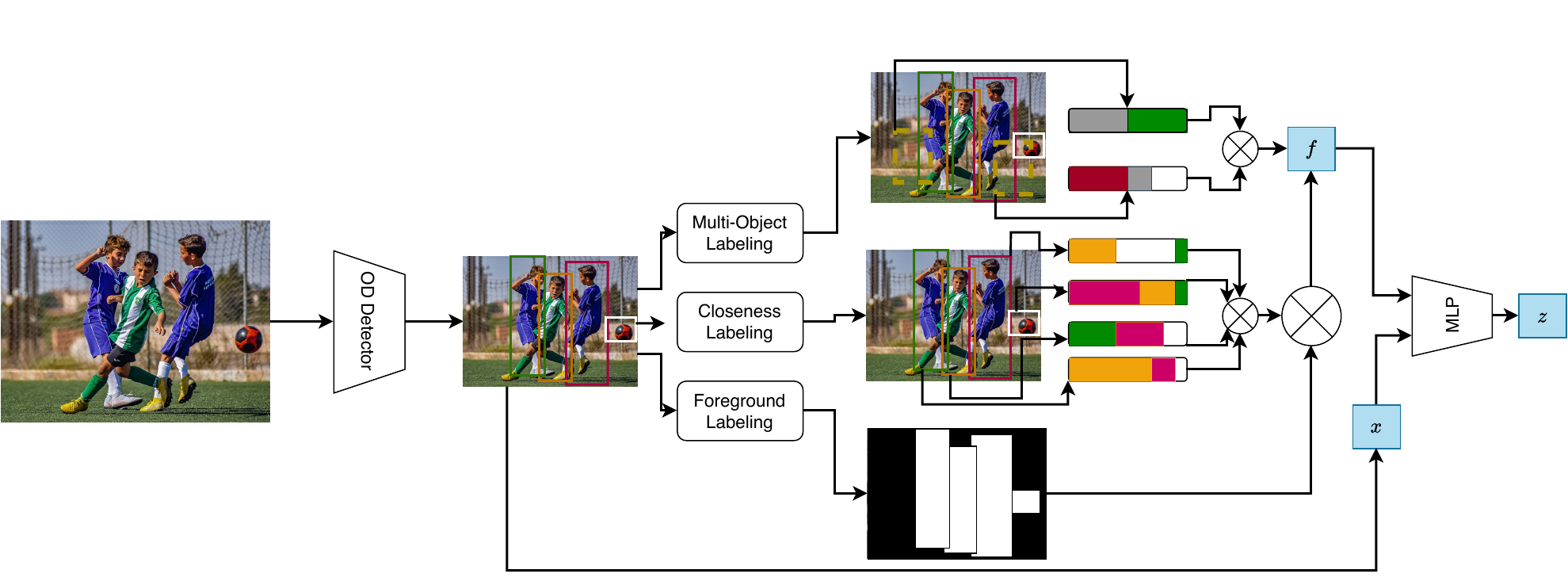}
\caption{Auxiliary Tasks implemented by \cite{recycling-bounding-box-annotations}. From top to bottom, \textit{Multi-object labeling}, \textit{Closeness labeling}, \textit{Foreground labeling}. 
Outputs are concatenated and the resulting representation $f$ is fed as input to a Multi-Layer Perceptron (MLP), along with the initial prediction $x$ for refinement.}
\label{recycling-auxiliary-tasks}
\end{figure*}

This capacity to leverage MTL to enhance the shared representation of tasks has motivated applications in diverse areas. For instance, \citet{self-supervised-monocular-road-segmentation} pre-train a CNN encoder on stereo-paired images from the well-known road object detection dataset KITTI \cite{KITTI} to perform monocular road segmentation. To achieve this, the authors choose to learn two tasks; \textit{Drivable Space Estimation} and \textit{Surface Normal Estimation}. Given a stereo-pair of images $(I_{left}, I_{right})$, the authors obtain a pseudo disparity map $I_{disparity}$ by using semi-global matching (SGM) \cite{SGM}. Subsequently, the authors run the Stixel World algorithm \cite{Stixel-world} which, given a RGB image $I_{RGB}$ ($I_{left}$ or $I_{right}$), exploits the corresponding disparity map $I_{disparity | I_{RGB}}$ to return a semantically segmented representation. Maximum a-posteriori (MAP) estimation is then performed based on the resulting distribution of the predicted pixel labels to extract the drivable area. Subsequently, surface normals are obtained by following the method introduced by \cite{surface-normal-unsupervised}. Specifically, given camera-related information such as the baseline distance $D$ and the focal length $D_{focal}$, the previously calculated diversity map $I_{disparity}$ is converted into a depth map $I_{depth}$. This depth map is later projected onto 3D world space $W$ given $D$ and $I_{normals}$ and is obtained via calculating the least-squares plane within $W$ and allocating the planes to neighbouring set of pixels. The authors fine-tune the learned features to perform monocular road segmentation and show impressive results whilst heavily reducing the demand for data.

\citet{recycling-bounding-box-annotations} utilise Self-MTL as a way to refine preliminary Object Detection (OD) predictions. In particular, assuming bounding box labels ${A_{OD}}$ are only available for object detection, 3 auxiliary tasks recycle ${A_{OD}}$ to produce their own respective labels ${A^{t}}$. Such a strategy has two main goals: (1) to learn robust discriminatory features for OD, (2) to refine the preliminary OD prediction. These auxiliary tasks are carefully chosen as follows: First, \textit{Multi-Object Labelling} randomly produces bounding boxes over the input image, constrained by the fact that one must overlap with at least one Ground Truth (GT) bounding box. Then, labels are assigned to the sampled bounding boxes based on GT Bounding Box area it overlaps the most with. The intuition behind this task is to perform augmentation on the input image to enhance globalisation. Second, \textit{Closeness Labeling} accounts for the inherent proximity in object classes in an image. This task consists in iterating over the GT bounding box annotations to provide a one-hot encoding based on the proximity of neighbouring GT bounding boxes. Finally, \textit{Foreground labeling} encodes the foreground and background, assigning 1's to pixels within GT bounding boxes and 0's otherwise. These tasks are illustrated in \cref{recycling-auxiliary-tasks}. Information encoded by these tasks is concatenated into a representation $f$ and is used to update the original prediction $x$ via a 1-layer FC layer to obtain a final refined prediction $z$ such that: $z = f \oplus x$. 

These methods demonstrate how effectively leveraging multiple self-supervised objectives can improve a shared representation suitable for MTL. Such efficiency has motivated some works to employ Self-MTL for diverse target downstream tasks in CV. For example, \citet{self-supervised-image-aesthetic-assessment} suggest a meaningful self-supervised pre-training strategy for Image Aesthetic Assessment (IAA). IAA models, which are usually trained an aesthetic-labeled ImageNet dataset \cite{ImageNet}, do not provide much information for why an image is not aesthetically good, for example, intrinsic image characteristics (\ie brightness, blurriness, contrast etc). Therefore, the authors train a comparative network of 2 distorted images, the distortion is chosen as one of the aforementioned characteristics and the networks aim at estimating the type of distortion as well as its intensity in an unsupervised manner. The goal of the MTL system is to recognise the less distorted image. Moreover, additional tasks are added to recognise the type and intensity of the distortion operation applied to the two input images. The authors report a decrease in 47\% in the number of epochs necessary for convergence compared to a IAA network pre-trained on Imagenet \cite{ImageNet}, notwithstanding the reduced need for data.

Alternatively, self-MTL framework has shown state-of-the-art  results in real-time applications. For example, SSMTL \cite{anomaly-detection} tackles anomaly detection in videos. Acquiring anomalous labels is difficult and as a result, the authors leverage self-supervised tasks to train a 3D CNN to recognise anomaly in videos. SSTML \cite{anomaly-detection} first runs a pre-trained YOLOv3 \cite{YOLOv3} to identify objects on a set of object-level frames ${I_{n}}$. Then, the authors choose three tasks to identify anomalous objects. First, irregularity is identified through the \textit{arrow of time} task, which involves obtaining an abnormal label by training the 3D CNN on the video in reverse mode. Second, \textit{motion irregularity detection} for which abnormal events are obtained via skipping frames is used to identify irregular motions such as someone running, falling etc. Third, a \textit{middle box prediction} task is implemented to predict the middle frame. Last, the authors enhance their multi-task 3D CNN through \textit{knowledge distillation} where the object detector YOLOv3  \cite{YOLOv3} is trained to predict the last layer of a ResNet-50 \cite{resnet}, which predicts whether the middle box frame is abnormal or not. The key point is that, in the knowledge distillation head, the authors expect a high difference between the object-level predictions of the 3D CNN and the ResNet-50 predictions when an anomaly is observed. The results significantly outperform previous state-of-the-art methods. Moreover, SSMMTL++ \cite{SSMTL++} recently reviews this framework and further improves it through the introduction of different tasks such as optical flow and advanced architectures such as the ViT \cite{ViT}. 

In addition to using multiple auxiliary tasks to enhance the learned representation, multiple modalities can be utilised to provide even more useful sources of information for models to learn. Multi-modal representation learning can be achieved by pre-training on diverse datasets. For instance, \citet{12-in-1} obtain a vision-language representation by pre-training on 12 vision-linguistic datasets and shows impressive results on common multi-modal tasks such as visual question answering and caption-based image retrieval. The authors utilise multi-modal self-supervision, inspired by \cite{Vilbert}, by masking proportional amounts of both image and word tokens and also by performing \textit{multi-modal alignment}, by predicting if two instances belong together. Similarly, \citet{sound-and-visual-rep-learning} introduce Multi-Self Supervised Learning tasks (Multi-SSL), a multi-modal (sound and image) pre-training strategy aiming to provide a shared representation for both sound and image modalities that could be used for downstream tasks. 

\begin{figure}[t!]
\centering
\includegraphics[width=8.5cm]{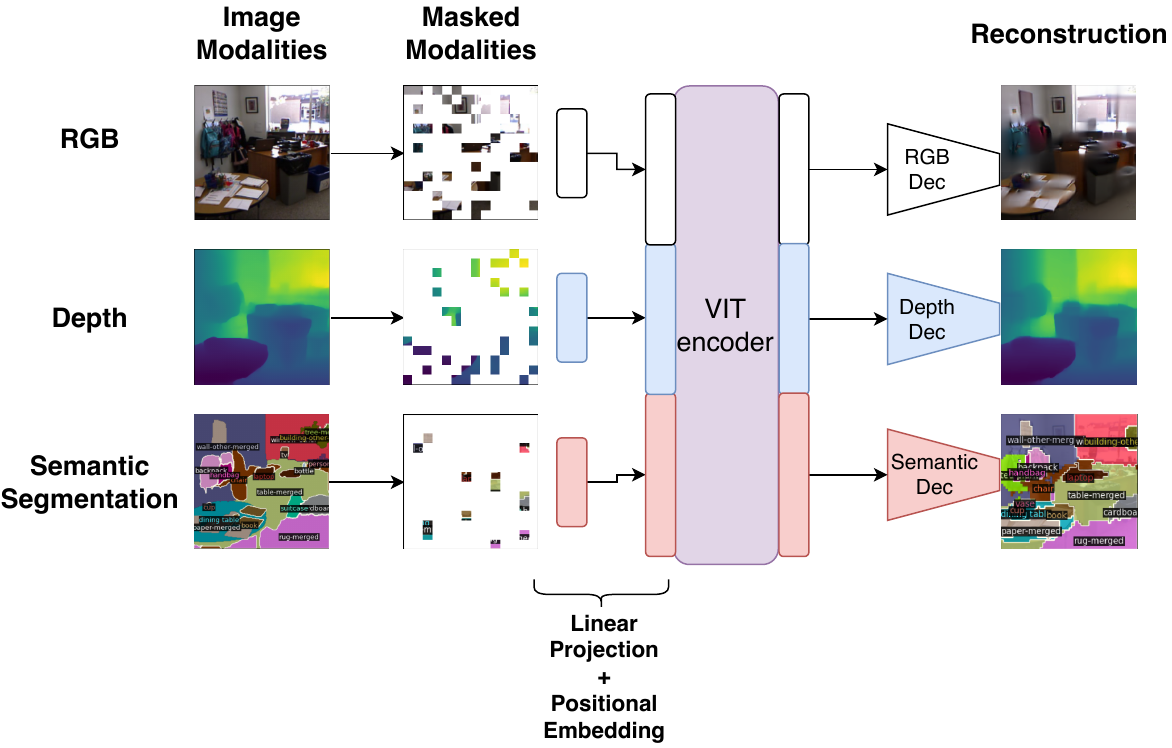}
\caption{\cite{MultiMAE} leverages three image modalities, obtained via pre-trained models, as a pre-training strategy. Respective image modalities are patch-wise masked in a similar way as MAE \cite{MAE}. Subsequently, linear projection and positional embedding are applied on patches. Then, patches are given as input to a shared ViT \cite{ViT} encoder which processes all the different representations. Finally, task-specific decoders aim at reconstructing each modality.}
\label{MultiMAE-figure}
\end{figure}

\citet{MultiMAE} leverage the recent the success of Masked Auto-Encoders (MAEs) \cite{MAE}. MAEs \cite{MAE} are asymmetric encoder-decoder models in which the encoder only operates on a small portion (about 15 \%) of a patch-wise masked input image and  the decoder aims at regenerating the missing patches. In particular, \citet{MultiMAE} propose Multi-Task MAE (MultiMAE), a pre-training strategy reconstructing diverse image modalities. To achieve this, given a set of RGB images, image modalities are acquired solely via \textit{Pseudo-labeling}. First, the depth modality is approximated by running a pre-trained DPT-Hybrid \cite{ViT-for-dense-prediction}, a ViT-based model. Similarly, Semantic Segmentation pseudo-labels are obtained via Mask2Former \cite{Mask2Former} trained on the COCO dataset \cite{COCO}. Once these labels are obtained, similar to original MAE \cite{MAE}, the authors sample a large portion of the image modalities divided into 16x16 patches. Subsequently, a number of tokens corresponding to approximately $\frac{1}{8}$ of the entire number of tokens for the 3 modalities (RGB, depth and semantic) are kept visible. The sampling strategy follows a symmetric Dirichlet distribution, equivalent to a uniform distribution so that no modality is prioritised. Then, the authors perform a 2D-sine-cosine linear embedding on the patches which are fed as input to the multimodal ViT encoder which operates only on the visible tokens, tremendously reducing the cost of computation \cite{MAE}. For downstream tasks, the multi-modal self-trained encoder can be used to fine tune a single task whilst benefiting from geometrical cues induced by other modalities. This framework is illustrated in \cref{MultiMAE-figure}.

In addition, Muli-Task Self-Supervised pre-training has been investigated in medical applications \cite{retinal-vessel-segmentation, MUSCLE, skin-lesion}, in music classification \cite{music-classification} or in NLP for multilingual reverse dictionaries \cite{multilingual-reserved-dictionary}.

\subsection{Semi-Supervised Learning Methods}
\label{sec:semi-supervised}
\subsubsection{Traditional Methods}

\citet{Semi-supervised-MTL} propose the first semi-supervised MTL framework. The framework consists of $T$ classifiers whose parameters share a joint probability distribution based on a soft variant of a Dirichlet Process. This allows for the parameters to be trained together and for the predictions to be obtained all at once. 
The probability distribution variant retains the inherent clustering property of Dirichlet Processes and as a result, the authors process unlabeled data via Parameterized Neighborhood-based Classification (PNBC). More specifically, the authors perform a Markov random walk over neighbouring data points obtained via supervised training, then, classifiers learn to assign unlabeled data to its closest point. 
Later, \citet{task-regularizations} expand on this setting by framing MTL as a clustering problem. To achieve this, after training $T$ linear classifiers, the authors improve their generalisation w.r.t. to their respective data by imposing a norm over the classification weights. Subsequently, the algorithm follows the same procedure, frames the respective classifiers into clusters via K-means clustering and assigns unlabeled points to nearby classifiers within that space. The authors also show this framework can be extended to non-linear classification through the use of kernels. 
It is worth noting that these traditional methods had a different notion of the MTL problem. In fact, the tasks are classification tasks in which `tasks' are either different datasets \cite{task-regularizations} or classes, resulting in multi-class classification \cite{Semi-supervised-MTL}. 
As a result, only one loss function is used for the optimisation which significantly differs from the contemporary definition of MTL.

\subsubsection{Self-Supervised-Semi-Supervised Methods}
The methods introduced in \cref{sec:representation-learning} highlight how multi-task learning can be used with self-supervised auxiliary tasks to minimise the overall training cost and demand for data. This characteristic has motivated numerous works to leverage both semi-supervised learning and self-supervised learning.

As explained in \cref{sec:task-grouping}, some tasks provide global understanding of scene geometry (\ie \textit{surface normals, depth prediction ...}) and when trained adequately, translate into low-level features tailored for dense prediction tasks. 
Therefore, there has been effort to investigate these tasks to improve an important CV task: \textit{Semantic Segmentation} (SS). For instance, \cite{urban-scene-understanding, dynamic-object-problem} use depth prediction as a proxy task for supervised urban scene understanding tasks such as car detection, road and semantic segmentation. Similarly, \citet{boostinc-semantic} use both depth estimation and colorization as a pre-training strategy for semantic segmentation in autonomous driving. 
To expand upon the idea that self-supervised depth estimation (SDE) can be effective to reduce data dependency, \citet{three-ways} introduce three ways to leverage SDE to improve semantic segmentation in a semi-supervised learning paradigm. 

First, the authors suggest an active learning strategy based on depth prediction. Specifically, given a set of images of the same domain ${G}$, the authors aim to split it into two image subsets. On the one hand, ${G_{A}} \subset {G}$ will be used for pseudo-labeled annotations for SDE, whilst ${G_{U}} \subset {G}$ is the set of unlabelled images. To obtain these, the authors iteratively choose ${G_{A}}$ through diversity sampling. Precisely, diversity is obtained when the chosen images are most representative of the dataset distribution. In urban scene understanding, this could result in the most frequent types of buildings, cars, bicycles, etc being chosen. To achieve diversity, the authors first populate ${G_{A}}$ with a random image $I$ from an image set $\{I\}$ and iteratively select the farthest $L_2$ distance between two sets of features of both ${G_{A}}$ and ${G_{U}}$ as given a pre-trained network $f_{SDE}$:
\begin{equation} \label{eq:active-learning}
G_{A_{n+1}} = \underset{I_{i} \in G_{U} }{\arg \max} \underset{I_{j} \in G_{A}}{\min} \lVert f_{SDE}(\theta, I_{i}) - f_{SDE}(\theta, I_{j})\rVert_{2},
\end{equation}
where the $f_{SDE}$ outputs the post-inference features based on the same set of input features $\theta$ and the respective annotated and unlabeled image sets $G_{A}$ and $G_{U}$.


Subsequently, the authors aim to incorporate another important aspect to this active sampling: \textit{Uncertainty Sampling} which consists in choosing samples that are hard to learn for the current state of the model: formally, instances in ${G_{U}}$ for which the model's decision is close to the decision boundary. To achieve this, a student model $f'_{SDE}(\theta, I)$ is trained on ${G_{A}}$. The authors then measure the disparity, on ${G_{U}}$, of both the predictions of $f_{SDE}$ and those of $f'_{SDE}$. Formally, the difference is calculated using the $L_1$ distance as:

\begin{equation} \label{eq:depthmmix}
E(i) = \lVert log(1+f_{SDE}(\theta, I)) - log(1+f'_{SDE}(\theta, I)) \rVert_{1}.
\end{equation}

The authors choose to use the $log$ regulator to avoid close-range objects dominating the disparity difference. 
Conceptually, sampling based on these two characteristics benefits from diversified, complex and representative instances which results in a decreased demand for data samples. 

Second, inspired by the success of pair-wise data augmentation in CV \cite{CutMix, ClassMix}, \citet{three-ways} introduce \textit{DepthMix} as a way to further reduce this labeling demand. In this method, considering 2 images $I_{source}$ and $I_{target}$, the goal is to learn a binary mask $M$ over $I_{source}$. Specifically, the positive values in $M$ represent regions to be copied over $I_{target}$. As a result, the augmented image $I_{augmented}$ is obtained as:
\begin{equation} \label{eq:goal-depthmix}
I_{augmented} = M \odot I_{source} + (1 - M) \odot I_{target},
\end{equation}
where $\odot$ is the element-wise product. In contrast to existing data augmentation methods, \citet{three-ways} leverage depth to avoid violating geometric semantic relationships between objects. For example, it is undesirable to have a distant object in $I_{source}$ to be copied onto the forefront of $I_{target}$, or worse, to result in geometrically implausible situations like a close-range motorbike copied on the top of a close-range car. To mitigate this problem, the authors use depth predictions for both images noted as $D_{source}$ and $D_{target}$. To achieve this, given a shared location $(x,y)$, $M$ is constrained to select only pixels for whose depth values are smaller on $I_{source}$ than on $I_{target}$. This process is demonstrated as follows:
\begin{equation} \label{eq:depthmix}
M(a,b) =\left\{
    \begin{array}{ll}
        1 & \mbox{if } D_{source}(a,b) <  D_{target}(a,b) + \epsilon\\
        0 & \mbox{otherwise}
    \end{array}
\right.
\end{equation}
where $\epsilon$ is a small noise value to avoid conflicts of objects that are the same depth plane on both images such as curb, road and sky.

The final component introduced in \cite{three-ways} is a semi-supervised MTL network to perform both Depth Estimation and Semantic Segmentation.
\begin{figure}[t!]
\centering
\includegraphics[width=8.5cm]{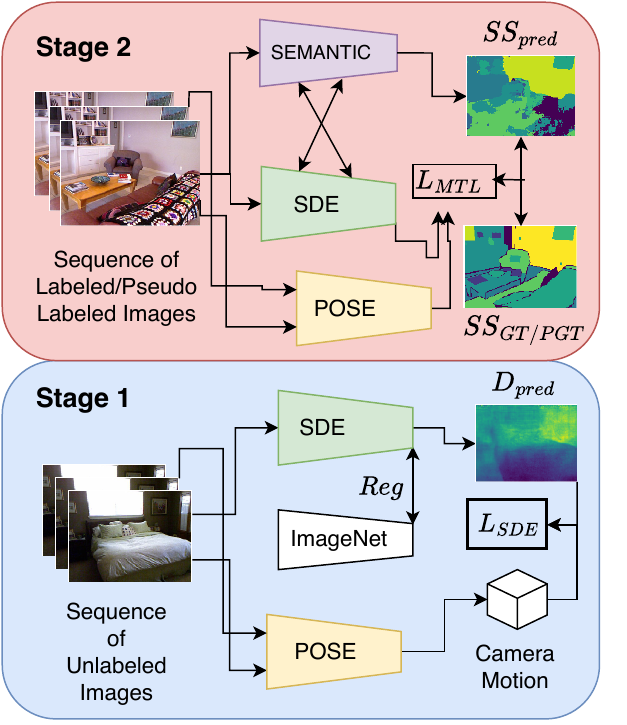}
\caption{\cite{three-ways} leverages Self-supervised Depth Estimation (SDE) to improve Semantic Segmentation. The training strategy is composed of two stages. In stage 1, a camera pose estimation network along with a SDE encoder is trained on an unlabeled sequence of images. Then, in stage 2, semantic segmentation is added and is trained on a sequence of images containing a mixture of labeled and pseudo-labeled images.}
\label{three-ways-MTL}
\end{figure}
The authors train their MTL network in 2 stages. The first stage is  depth pre-training. This stage consists in a self-supervised training for both depth estimation and pose estimation on an unlabeled sequence of images. As part of this procedure, a shared encoder $f_{\theta}^{E}$ is initialised with ImageNet \cite{ImageNet}. Additionally, in order not to forget the semantic features during training, the initialised features, noted as $f_{I}^{E}$, serve as a regulator for the SDE pre-training and the authors use the $L_2$-norm in order to guide the multi-task representation. The resulting loss term is formulated as:
\begin{equation} \label{eq:three-ways-regulation}
L_{SDE} = \lVert f_{\theta}^{E} - f_{I}^{E} \rVert_{2}.
\end{equation}
In the second stage, the authors introduce semantic segmentation to form a semi-supervised network. In this stage, the network is trained on depth estimation on both labeled and pseudo-labeled (using the mean teacher algorithm \cite{mean-teacher-algorithm}) instances. Their solution is illustrated in \cref{three-ways-MTL}.
As a result, the authors manage to achieve 92\% accuracy on a baseline fully-supervised model whilst using 1/30 of labeled image segmentation instances. Furthermore, whilst using 1/8 of the SS labels, it outperforms this supervised baseline by a small margin. The authors then improve their solution to perform domain adaptation \cite{semi-supervised-domain-adaptive}. 

Recently, \citet{MTL-for-image-segmentation-task} leverage both depth and surface normals estimation to improve on semantic segmentation. In addition, the authors show how Nash-MTL \cite{bargaining-game} can lead to efficient solutions.

\subsubsection{Generative Modeling}
Recent advances of general self-supervised methods such as adversarial training with Generative Adversarial Networks (GANs) \cite{GANs}, as well as the ability of generative modeling to learn useful visual representations from unlabeled images \cite{adversarial-feature-learning}, have motivated the investigation of generative modeling in MTL to lower the demand for labeled data \cite{adver-learning-semantic-semi-supervised, robust-adver-learning-for-semantic-semi}. 

\begin{figure}[t!]
\centering
\includegraphics[width=8.5cm]{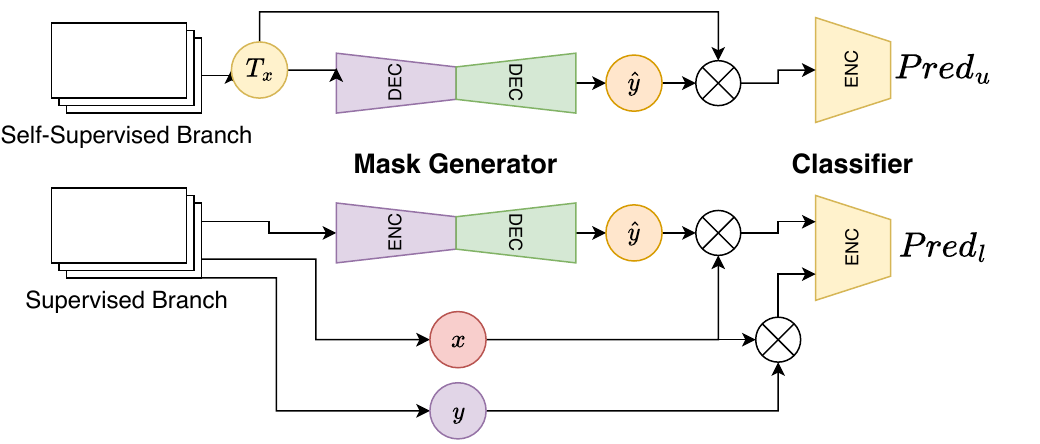}
\caption{In \cite{partly-supervised-MTL}'s solution, a two-branch MTL network shares the same Generator and Discriminator trained in an adversarial fashion. For the self-supervised branch, a transformation $T_{x}$ is applied to the input image.}
\label{S4MTL}
\end{figure}

For example, \citet{partly-supervised-MTL} propose a self-supervised semi-supervised MTL ($S^{4}MTL$) solution leveraging adversarial learning and semi-supervision to teach simultaneously two commonly tackled CV tasks, namely: Image Classification (for diagnostic classification) and Semantic Segmentation. By considering two datasets, one labeled $D_{A}$ and one unlabeled $D_{U}$, the authors define their respective losses as $L_{A}$ and $L_{U}$. If $\theta$ and $\upsilon$ define the parameters of network $f$ for semantic segmentation and diagnostic classification respectively, then the overall objective can be summarised as:
\begin{equation} \label{eq:partly-supervised-MTL}
\min_{\upsilon, \theta} L_{A}(D_{A},f(\upsilon, \theta)) + \alpha L_{U}(D_{U},f(\upsilon, \theta)),
\end{equation}
where $\alpha$ is a positive weight for the unsupervised loss. 
Subsequently, the authors train two networks: $G$, a mask generator for semantic segmentation and $D$ a classifier which is trained in an adversarial fashion. These two networks are divided into two branches. 
For supervised images, $G$ wants $D$ to maximise the likelihood of the segmentation masks given a regular image-label pair.
For the  unsupervised images, the model performs a transformation $t(x)$ over the input image $x$ such as rotation to enable $G$ to make predictions. Such a framework is illustrated in \cref{S4MTL}. 
Using this framework, the model is claimed to outperform fully-supervised single task models whilst diminishing the availability of data/label up to 50\%.

\citet{semi-supervised-MTL-for-semantics-depth} extend on this framework and introduces SemiMTL. This method performs urban scene segmentation and depth estimation. However, the authors leverage multiple datasets in a heterogeneous (trained on different datasets) MTL framework and train their discriminator $D$ in a domain-aware fashion to compensate for the domain shift inherent to this environment. To do this, the authors add a inter-domain loss between the labeled dataset $A$ and unlabeled dataset $B$ for which the ground-truth value for an arbitrary task $t$ is noted as $y^{A}_{t}$ and $y^{B}_{t}$. Moreover, their respective predictions are noted $\hat{y}^{A}_{t}$ and $\hat{y}^{B}_{t}$. The authors choose to leverage the cross-entropy loss and as a result, this inter-domain loss can be expressed, over the data instances $i$, as:
\begin{equation} \label{eq:inter-domain-loss}
L_{inter}^{t} = - \sum_{i}^{T} \log (D_{t}(\hat{y}^{B}_{t})^{(i,y^{A}_{t})}),
\end{equation}
where $y^{A}_{t}$ is a 3-dimensional one-hot vector, in which
a three-way classifier is utilized in the discriminator to tell that the input is from the ground-truth from dataset $A$.
Conceptually, the loss in \cref{eq:inter-domain-loss} aligns the unlabelled task prediction $\hat{y}^{B}_{t}$ onto the labelled task ground-truth $y^{A}_{t}$ to compensate for domain shifts. 
Additionally, the authors introduce different ground-truth and prediction alignment strategies such as aligning the unlabelled prediction $\hat{y}^{B}_{t}$ onto the labelled task prediction $\hat{y}^{A}_{t}$ or aligning $\hat{y}^{B}_{t}$ onto the intersection of the labelled ground-truth $y^{A}_{t}$ and prediction $\hat{y}^{A}_{t}$.

\subsubsection{Discriminative Methods}
Discriminative methods aim at determining boundaries between image representations by directly comparing them.
This section focuses on MTL works introducing this technique under semi-supervised training paradigms. 
One type of discriminative method that has shown great success in many CV tasks is \textit{Contrastive Learning} (CL). CL was  originally introduced by \cite{original-contrastive}. It involves learning a joint-space in which similar pairs of images are close to each other and in which different pairs are far part. Momentum Contrast (MoCo) \cite{Moco} extends this concept for unsupervised visual learning and sees this framework as a dictionary look-up problem where an image $I$ is encoded by a network $f$, this is denoted as the query $q = f(I)$. Then, a queue of size $n$ of image representations ${I_{k}}$, or keys, chosen as the preceding mini-batch, which are encoded by a momentum encoder $f_{m}$ are compared $k_{n} = f_{m}({I_{k}})$. Subsequently, the matching key $k_{+}$ is noise-augmented. Finally, $f$ is updated via the InfoNCE \cite{InfoNCE} as follows:
\begin{equation} \label{eq:infoNCE}
L_{InfoNCE} = - \log \frac{\exp(q \cdot k_{+} / \tau)}{\sum_{i=1}^{N} \exp(q \cdot k_{i} / \tau)}.
\end{equation}
SimCLR \cite{SimCLR} suggests a simpler version comparing diverse augmented versions of the same image, however it requires larger batch sizes.

\begin{table*}
\centering
\caption{Single-Task vs Multi-Task Fully-Supervised Methods Comparison on NYUv2}
\label{tab.1}
  \begin{tabular}{lllcc c c c}
    \toprule
    \multirow{2}{*}{Dataset} &
    \multirow{2}{*}{Method} &
    \multirow{2}{*}{MTL} &
      \multicolumn{1}{c}{Semseg} &&
      \multicolumn{1}{c}{Depth} &&
      \multicolumn{1}{c}{Normal}  
      \\
      \cline{4-4} \cline{6-6} \cline{8-8} \\
      && &  {mIoU $\uparrow$} && {RMSE $\downarrow$} && {mErr $\downarrow$} \\ 
      \midrule
    \multirow{20}{*}{NYUv2 \cite{NYUv2}}
    & \citet{8578788} & \ding{55} & 48.10 && - && - \\
    & \citet{yu2020multilayer} & \ding{55} & 50.70 && - && - \\
    & InverseForm \cite{borse2021inverseform} & \ding{55} & 53.10 && - && - \\
    & TADP \cite{unidepth} & \ding{55} & - && 0.225 && - \\
    & DepthAnything \cite{depth-anything} & \ding{55} & - && 0.206 && - \\
    & UniDepth \cite{unidepth} & \ding{55} & - && 0.201 && - \\
    & \citet{hickson2019floors} & \ding{55} & - && - && 19.7 \\
    & \citet{bae2021estimating} & \ding{55} & - && - && 14.9 \\
    & iDisc\cite{piccinelli2023idisc} & \ding{55} & - && - && 14.6 \\
    \\
    \cline{2-8}
    \\
    & Cross-Stitch \cite{cross-stich} & \ding{51} & 36.34 && 0.6290 && 20.88\\
    & PAP \cite{PAP} & \ding{51} & 36.72 && 0.6178 && 20.82\\
    & PSD \cite{PSD} & \ding{51} & 36.69 && 0.6246 && 20.87\\
    & PAD-Net \cite{PAD-net} & \ding{51} & 36.61 && 0.6270 && 20.85\\
    & MTI-Net \cite{MTI-NET} & \ding{51} & 45.97 && 0.5365 && 20.27\\
    & InvPT \cite{invPT} & \ding{51} & 53.56 && 0.5183 && 19.04\\
    & TaskPrompter \cite{taskprompter} & \ding{51} & 55.30 && 0.5152 && 18.47\\
    & DeMT \cite{DeMT} & \ding{51} & 51.50 && 0.5474 && 20.02\\
    \bottomrule
  \end{tabular}
\end{table*}

Motivated by the aforementioned approaches, MTSS \cite{MTSS} suggests a simple, yet effective, semi-supervised MTL framework to optimise a discriminative self-supervised auxiliary task and a supervised main task simultaneously.
\begin{figure}[t!]
\centering
\includegraphics[width=8.5cm]{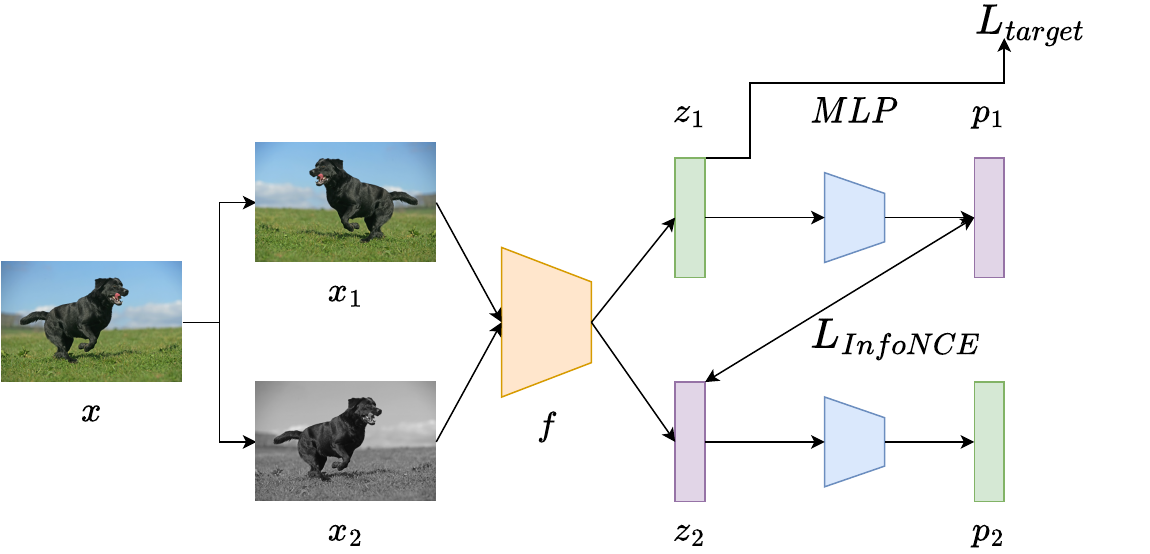}
\caption{\cite{MTSS} suggests to perform two augmentations of the same image. Each of the representations are encoded through the CNN encoder $f$, resulting in representations $z_{1,2}$. Whilst one representation (i.e $z_{1}$) can be processed by the main task, the InfoNCE \cite{InfoNCE} loss is minimised between $z_{2}$ and the encoded representation $p_{1}$.}
\label{MTSS-figure}
\end{figure}
Specifically, the authors choose to maximise the similarity between two different views of the same image. First, two augmentations on the same image are performed, these views are $x_{1}$ and $x_{2}$. Then, a shared CNN classifier process them leading to two representations $z_{1}$ and $z_{2}$. One, for example $z_{1}$, is chosen to be processed by the supervised main task. Similarly to SimCLR \cite{SimCLR}, the authors choose to attach a Multi-Layer Perceptron (MLP) in order to map representations to a similar space, let us denote the resulting representations as $p_{1}$ or $p_{2}$. Finally, the cosine similarity $D$ between $p_{1}$ and $z_{2}$ is calculated as shown below: \begin{equation} \label{eq:similarity-MTSS}
D(p_{1}, z_{2}) = - \frac{p_{1}}{\lVert p_{1} \rVert_{2}} \cdot \frac{z_{2}}{\lVert z_{2} \rVert_{2}},
\end{equation}
to minimise the cosine similarity between the representations of augmented views. The symmetric auxiliary loss, introduced by BYOL \cite{BYOL} and depicted in \cref{eq:BYOL}, is used as follows:
\begin{equation} \label{eq:BYOL}
L_{aux} = \frac{1}{2} D(p_{1}, z_{2}) + \frac{1}{2} D(p_{2}, z_{1}).
\end{equation}
This auxiliary loss is then added to the overall MTL objective. The semi-supervised  framework is depicted in \cref{MTSS-figure}.

\begin{figure*}[t!]
\centering
\includegraphics[width=0.95\textwidth]{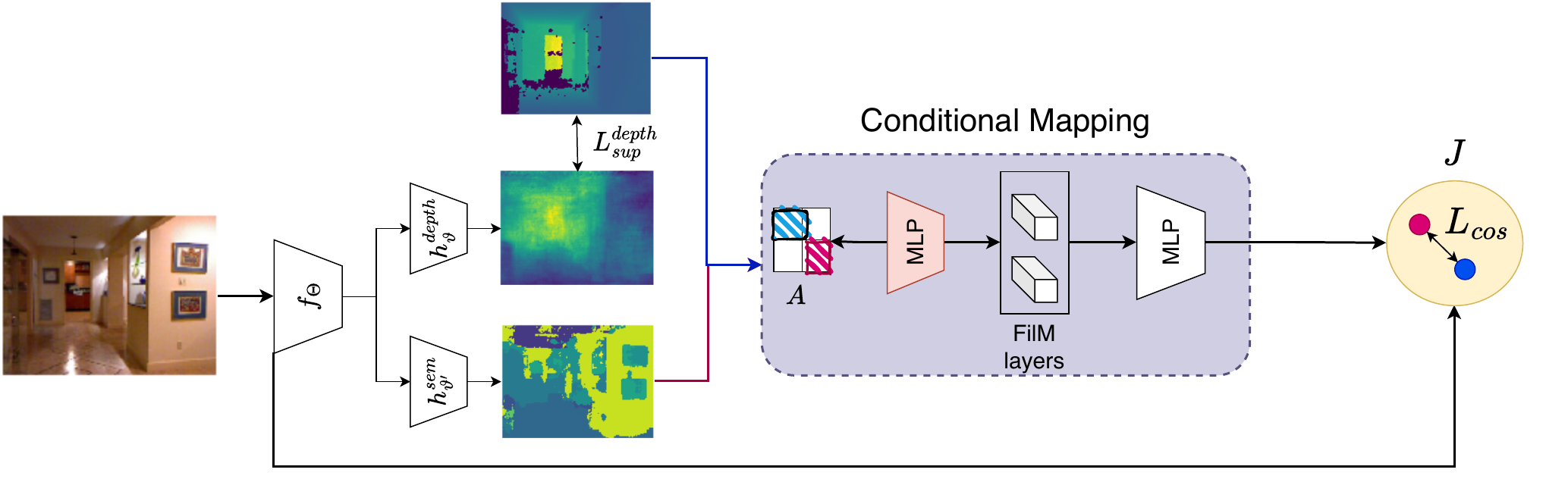}
\caption{Considering an input image for which only the depth ground truth is available, \cite{MTPSL} performs cross-task consistency and maps the depth ground-truth to the semantic segmentation prediction to a joint space $J$ through a conditional mapping network (in purple). The cosine distance between the two representations is minimised.}
\label{MTPSL-fig}
\end{figure*}

Another recent task discriminatory approach, Cross-Task Consistency (XTC), is introduced by \cite{X-task-consistency}. Conceptually, this notion comes from the dependency between two tasks. For instance, in the context of urban scene semantic segmentation with depth estimation, there would be inconsistency if depth estimation evaluated a flat surface where a car is detected. Therefore, \citet{X-task-consistency} aim to compute task pair-wise mapping to map the prediction from a source task to the label of the target task. 
However, each of those mapping functions are parameterised by two Deep Neural Networks (DNNs) and leverage labels from each task. To mitigate the use of labeled data, \citet{MTPSL} leverage cross-task relations in a semi-supervised framework. Specifically, \cite{MTPSL} suggests a framework to map the prediction of an unlabeled task $\hat{y}^{s}$ to the ground truth of another task $y^{t}$ through an adaptive encoder which embeds only shared parameters. Therefore, the two representations $\hat{y}^{s}$ and $y^{t}$ are mapped on to a joint space and their  cosine distance is minimised.

\citet{MTPSL} leverage XTC in their framework for semantic segmentation, depth estimation and surface normals estimation. Let us consider a partially-supervised image $I$, for which only $y^{depth}$ or $y^{semantic}$ is available. $I$ is then processed through a shared backbone network $f_{\Theta}$ to which task-specific decoders $h_{\vartheta}^{depth}$ and $h_{\vartheta'}^{semantic}$ are attached. 
The obtained predictions are noted as $\hat{y}^{depth}$ and $\hat{y}^{semantic}$. For the sake of illustration, let us consider 
$\hat{y}^{semantic}$ not to be labelled and therefore to leverage the available ground-truth from the depth estimation task. Now describing the XTC mechanism, let us consider a matrix $A$ for which entries correspond to $source \rightarrow target$, (in our example, $A[semantic,depth] = 1$) and all other entries are 0. An auxiliary network $k_{\theta}$ is used to conditionally parameterise a mapping network $m_{\psi}$. Similar to \cite{film}, $k_{\theta}$ is used to update the layers of $m_{\psi}$. This mechanism is to allow for a conditional source-to-target mapping. The two resulting representations are then projected on to the same joint-space $J$. The authors use the cosine similarity to minimise their distance. 
Additionally, to avoid trivial mappings, the features from $f_{\Theta}$ are used as a regularisation term of the distance between the mapping function's output and the encoded features $f_{\Theta}(I)$. The explained mapping is illustrated in \cref{MTPSL-fig}.

\subsection{Few-Shot Learning Methods}
\label{sec:few-shot-learning}
Few-Shot Learning (FSL) is a learning paradigm that aims to learn unseen classes from a few examples. This training paradigm is motivated by the fact that humans do not need hundreds or thousands of exemplar images to learn to recognise an object. Typically, FSL systems consist of two stages. First, a general feature extractor is learned from a large annotated dataset in a stage called \textit{meta-training}. Second, an adaption strategy is used to classify the new sample/class (also known as the query sample) based on a small labeled support set. This stage is called \textit{meta-testing}. A similarity function is then used on the support set to identify the matching class given the query sample. 
Traditionally, in FSL-MTL, the goal is to adapt to unseen classes for a specific task within a MTL model. In the context of MTL, cross-task interactions within a multi-task system could help enhance the generalisation to the few-shot target task. In fact, \citet{natural-language-decathlon} show that MTL models generally focus on tasks that have the least training samples, which is due to the feature sharing process across tasks. 

Recently, the FSL literature has heavily focused on the initial meta-training stage in which multiple datasets serve to train a model to obtain global representations for a target few-shot task, most commonly being \textit{image classification}. For example, \citet{improving-fsl-with-self-pretext-tasks} suggest training such a model in a MTL fashion by leveraging self-supervised tasks (similar to solutions introduced in \cref{sec:representation-learning}), on both labelled and unlabelled images. The shared encoder is regularised by the contrastive learning method: BYOL \cite{BYOL}. Subsequently, the MTL system is evaluated on traditional few-shot image classification. 

MTFormer \cite{MTFormer} suggests different dense prediction tasks as few-shot tasks and evaluates a MTL system leveraging a cross-task attention mechanism at the decoder level of a ViT \cite{ViT} on the PASCAL dataset \cite{PASCAL}. The authors evaluate three tasks, in turn, as a few-shot sampled task by randomly sampling about 1 \% of the annotated data for the few-shot task and keeping all available labels for other tasks. MTFormer \cite{MTFormer} chooses to evaluate Semantic Segmentation, Human Part Segmentation and Saliency Detection which consists of identifying interesting points in an image (points that the human eye would focus on straight away). The results, presented in \cref{tab.5}, display an impressive improvement over the single-task FSL baseline. This improvement is explained by two techniques: the feature propagation across tasks to enhance the few-shot task representation, and the use of CL in \cite{MTFormer}, in which different task representations of the same image are considered as positive samples, which further reinforces the shared representation's quality. 

\textit{Visual Token Matching} (VTM) \cite{visual-token-matching} proposes a continual few-shot learning framework for dense prediction vision tasks. In this setting, a universal few-shot learner can learn new dense prediction tasks given extremely limited labelled task images, most often only using 10 labelled examples of image-label pairs. VTM employs a encoder-decoder architecture using ViT encoders \cite{ViT} to encode both image and label. As a way to propagate features across the model hierarchies, the authors perform token matching using an attention mechanism similar to MTFormer \cite{MTFormer}. More specifically, given a target few-shot task $t$, a query image $Q_{t}$ and support set of image-label pairs of length $N$ ($(X,Y)^{1...N}_{t}$), a task-specific shared encoder $f_{t}$ is used to process both $Q_{t}$ and $X^{i}_{t}$. On the other hand, a label encoder $g$ is used to encode $Y^{i}_{t}$. Subsequently, the token matching mechanism based on attention operates on ViT blocks representations. The block-wise query label predictions are then concatenated before a classification head provides the final prediction. Finally, the results reported by \cite{visual-token-matching} suggest similar strategies should be elevated to the simultaneous MTL settings.

\begin{table}[h]
\centering
\caption{Fully-Supervised MTL methods on PASCAL-Context}
\label{tab.2}
\begin{tabular}{cccc}
\toprule
\multirow{2}{*}{\textbf{Model}} & \textbf{Semseg} & \textbf{Parsing} &  \textbf{Saliency}  \\
& mIoU $\uparrow$ & mIoU $\uparrow$ & maxF $\uparrow$  \\
\midrule
Cross-Stitch \cite{cross-stich} & 63.28 & 60.21 & 65.13\\
PAD-Net \cite{PAD-net} & 60.12 & 60.70 & 67.20\\
MTI-Net \cite{MTI-NET} & 61.70 & 60.18 & 84.78\\
InvPT \cite{invPT} & 79.03 & 67.71 & 84.81\\
MTFormer \cite{MTFormer} & 74.15 & 64.89 & 67.71\\
TaskPrompter \cite{taskprompter} & 80.89 & 68.89 & 84.83\\
DeMT \cite{DeMT} & 75.33 & 63.11 & 83.42\\
\bottomrule

\end{tabular}
\label{tab:example_multirow}
\end{table}

\begin{table*}
\centering
\caption{Semi-Supervised Learning (MTPSL \cite{MTPSL}) Comparison on NYUv2 and Cityscapes}
\label{tab.3}
  \begin{tabular}{lllcc c c c} 
    \toprule
    \multirow{2}{*}{Dataset} &
    \multirow{2}{*}{Method} &
      \multicolumn{1}{c}{Semseg} &&
      \multicolumn{1}{c}{Depth} &&
      \multicolumn{1}{c}{Normal} & 
      \\
      \cline{3-3} \cline{5-5} \cline{7-7} \\
      && {mIoU $\uparrow$} && {aErr $\downarrow$} && {mErr $\downarrow$} \\ 
      \midrule
    \multirow{ 6}{*}{NYUv2 \cite{NYUv2}}
    & $STL_{SS}$ & 37.45 && - && - \\
    & $STL_{Depth}$ & - && 0.61 && - \\
    & $STL_{SN}$ & - && -  && 25.94 \\
    & $MTL_{CNN}$ & 36.95 && 0.55 && 29.5 \\
    \\
    \cline{2-8}
    \\
    & \cite{MTPSL} MTPSL (1/3) & 28.43 && 0.63 && 33.01 \\
    & \cite{MTPSL} MTPSL (one) & 31.00 && 0.51 && 28.58 \\
    \midrule
    \multirow{ 8}{*}{Cityscapes \cite{cityscapes}}
    & $STL_{Seg}$ & 74.19 && -  \\
    & $STL^{SegNet}_{Depth}$ &  - && 0.012 \\
    & $MTL_{CNN}$ & 73.36 && 0.016  \\
    \\
    \cline{2-8}
    \\
    & \cite{MTPSL} MTPSL (one) & 74.90 && 0.016 \\
    & \cite{MTPSL} MTPSL (1:9) & 71.89 && 0.013 \\
    & \cite{MTPSL} MTPSL (9:1) & 74.23 && 0.026 \\
    \bottomrule
  \end{tabular}
\end{table*}

\begin{table*}
\centering
\caption{Semi-Supervised Learning (MTPSL \cite{MTPSL}) on PASCAL-Context}
\label{tab.4}
  \begin{tabular}{lllcc c c c c c c}
    \toprule
    \multirow{2}{*}{Dataset} &
    \multirow{2}{*}{Method} &
      \multicolumn{1}{c}{SemSeg} &&
      \multicolumn{1}{c}{Human Parts} &&
      \multicolumn{1}{c}{Normal} && 
      \multicolumn{1}{c}{Saliency} &&
      \multicolumn{1}{c}{Edge} 
      \\
      \cline{3-3} \cline{5-5} \cline{7-7} \cline{9-9} \cline{11-11}\\
      && {mIoU $\uparrow$} && {mIoU $\uparrow$} && {mErr $\downarrow$} && {mIoU $\downarrow$} && {odsF $\uparrow$} \\ 
      \midrule
    \multirow{ 4}{*}{Pascal-Context \cite{PASCAL}}
    & STL & 47.7 && 56.2 && 16.0 && 61.9 && 64.0  \\ 
    \\
    \cline{2-11}
    \\
    & \cite{MTPSL} MTPSL (one) & \textbf{49.5} && 55.8 && 17.0 && 61.7 && \textbf{65.1}  \\ 
    \bottomrule
  \end{tabular}
\end{table*}

\section{Datasets \& Tools}

\cref{sec:datasets} refers the reader to a list of datasets commonly utilised in MTL for computer vision. Additionally, \cref{sec:discussion} provides a summary of the results achieved by partially-supervised MTL solutions. Based on these results, we discuss and analyse common trends and suggest interesting paths of exploration to further improve MTL. Last, we introduce a table summarising the different open-source MTL code.

\subsection{Datasets}
\label{sec:datasets}
Below is a list of common multi-task CV datasets.
\begin{enumerate}
  \item \textbf{Taskonomy. }\cite{taskonomy} This dataset is the largest multi-task dataset. It contains 4.5 million indoor scene images, each labeled with 25 annotations. These images include: scene annotations, camera information, 2D/3D keypoints, surface normals and various-level object annotations. The foundational work \cite{taskonomy} on this dataset performed experiments on 26 diverse tasks.
  
  \item \textbf{NYUv2-Depth. }\cite{NYUv2} This dataset comprises 1449 labeled images drawn from indoor scene videos for which each pixel is annotated with a depth value and an object class. Additionally, there are 407,024 unlabeled images which contain RGB, depth and accelerometer data, rendering this dataset useful for real-time applications as well.  

  \item \textbf{Cityscapes. }\cite{cityscapes} This dataset consists of 5000 urban scenes. Each image is annotated with pixel-level labels for 30 classes. Additionally, the dataset includes image stereo pairs associated camera shift metadata. Therefore, \cite{cityscapes} leverages stereo-paired information to produce accurate depth labels. As a result, Cityscapes \cite{cityscapes} is typically used as a 7-class semantic segmentation class and depth estimation task. 

  \item \textbf{Pascal-Context. }\cite{PASCAL} A dataset of 1464 of regular object centered scenes. This dataset includes tasks such as  saliency estimation, depth estimation, human part segmentation as well as semantic segmentation.

  \item \textbf{KITTI. }\cite{KITTI} This dataset is one of the most popular datasets for Autonomous Driving. The images result from hours of driving in diverse traffic environments. This dataset has been utilised for 3-class \cite{3-class-kitti}, 10-class \cite{10-class-kitti} or 11-class \cite{11-class-kitti} semantic segmentation or object detection. Additionally, the dataset includes 3D labeled point clouds for 15,000 images.  
\end{enumerate}

\subsection{Results and Discussion}
\label{sec:discussion}
This section presents results for partially supervised MTL. Moreover, an attempt to derive both general performance guidelines and future areas of investigation is made. 

\cref{tab.1} provides a comparison of traditional single-task methods with a range of recent multi-task learning methods. The single-task methods covered in this table use RGB-only processing to provide a fair comparison. This table reviews three traditionally tackled tasks : semantic segmentation, monocular depth estimation and surface normal estimation. By analysing the presented methods on the NYUv2 Dataset \cite{NYUv2}, we can observe that semantic segmentation generally improves by taking advantage of the depth and surface normal features from  depth and surface. However, we can notice that, typically, MTL methods fail to perform as good as single-task methods on tasks like depth and surface estimation. We hypothesise that the reasons being (1) due to task-optimal network architectures not being the same for all the tasks, leading to a non-conceivable or overly complex MTL architecture; (2) a task-specific loss function designed generalising poorly to the MTL aggregated gradient representation and (3) the trend to design scalable and simple MTL networks with lightweight decoders which does not reflect well the difficulty of each task. 

Furthermore, we provide, in \cref{tab.2}, a summary of fully-supervised performant MTL methods on the Pascal-Context dataset \cite{PASCAL} covering commonly tackled tasks : semantic segmentation, human part parsing (which is semantic segmentation on human body parts) and saliency detection which consists of identifying interesting points in an image (points that the human eye would focus on straight away). We identify that a comparison with STL methods is complex due to the lack of STL methods covering the same split of the PASCAL dataset \cite{PASCAL}. We however notice a significant improvement brought by various MTL methods on the semantic segmentation: where the best STL method achieves 71\% mIoU \cite{hong2023minimalist}, 4 MTL methods significantly outperforms this result in \cref{tab.2} whilst performing human parsing and saliency detection. 

\begin{table*} 
\centering
\caption{MTFormer\cite{MTFormer} treats a \textcolor{red}{target task} annotations as few-shot samples whilst keeping two other tasks fully-supervised. \\
Results are reported on the PASCAL dataset \cite{PASCAL}.}
\label{tab.5}
  \begin{tabular}{lllllc c c c c}
    \toprule
    \multirow{2}{*}{Method} &&
    \multirow{2}{*}{Few-Shot Task} &&
    \multirow{2}{*}{SS $\uparrow$} &&
    \multirow{2}{*}{Human Part Seg. $\uparrow$} &&
    \multirow{2}{*}{Saliency $\uparrow$} \\
    \\
    \cline{5-5} \cline{7-7} \cline{9-9}
    &&&& {mIoU $\uparrow$} && {mIoU $\uparrow$} && {mIoU $\uparrow$} \\
    \midrule
      STL && SS && \textcolor{red}{3.34} && 63.90 && 66.71 \\
      MTFormer \cite{MTFormer} && SS && \textcolor{red}{35.26} && 64.26 && 67.26 \\ 
    \midrule
      STL && Human Part Seg. && 71.17 && \textcolor{red}{11.27} && 66.71 \\
      MTFormer \cite{MTFormer} && Human Part Seg. && 73.36 && \textcolor{red}{51.74} && 67.74 \\ 
    \midrule
      STL && Saliency && 71.17 && 63.90 && \textcolor{red}{44.39} \\
      MTFormer \cite{MTFormer} && Saliency && 76.00 && 66.89 && \textcolor{red}{55.55} \\ 
    \bottomrule
  \end{tabular}
\end{table*}

\begin{table*} 
\centering
\caption{MTL open-source code repositories }
\label{tab.6}
  \begin{tabular}{c c p{9cm}}
    \toprule
    \multicolumn{1}{c}{Type} &
    \multicolumn{1}{c}{Link} & 
    \multicolumn{1}{c}{Description}  \\
    \midrule
    \multirow{ 2}{*}{Paper Repository} 
    & \href{https://github.com/WeiHongLee/Awesome-Multi-Task-Learning}{Awesome Multi-Task Learning 1} & This repository regroups MTL-related papers in a chronological order.\\
    & \href{https://github.com/Manchery/awesome-multi-task-learning}{Awesome Multi-Task Learning 2} & This repository gathers MTL papers and provides a categorisation. \\
    \midrule
    
    \multirow{ 2}{*}{Programming Framework} 
    & \href{https://github.com/zhanglijun95/AutoMTL}{AutoMTL} \cite{AutoMTL} & This solution performs automatic MTL model compression given an arbitrary backbone and a set of tasks. \\
    &
    \href{https://github.com/median-research-group/LibMTL}{LibMTL} \cite{LibMTL} & This is a Python library for MTL built on Pytorch. The implementation supports a large number of SOTA solutions, weighting strategies and data loaders. \\
    \midrule
    \multirow{ 3}{*}{Benchmarking} 
    & \href{https://github.com/SimonVandenhende/Multi-Task-Learning-PyTorch}{Dense Prediction Tasks} \cite{MTL-dense-predic-tasks} & This solution benchmarks a 2 MTL solutions on CV dense prediction tasks on 2 datasets. It is implemented in Pytorch.\\
    & 
    \href{http://taskonomy.stanford.edu}{Taskonomy} \cite{taskonomy} & In addition to providing web-based visualisations. Taskonomy \cite{taskonomy} introduces a API to group 25 vision tasks. Pre-trained models are available in Tensorflow and Pytorch. \\ 
    &
    \href{https://github.com/SamsungLabs/MTL}{Aligned-MTL} & A programming repository introducing a new gradient-based optimisation technique and allowing to benchmark a wide range of different MTL optimisation strategies introduced in \cref{chapter:Optimisation}. \\ 
    \midrule
    \multirow{ 1}{*}{Self/Semi-supervision} 
    & \href{https://github.com/VICO-UoE/MTPSL}{MTPSL} \cite{MTPSL} & This solution implements different cross-task mapping under balanced and imbalanced semi-supervised settings for dense prediction tasks. This solution is implemented in Pytorch and supports two datasets.\\
     & \href{https://multimae.epfl.ch}{MultiMAE} \cite{MultiMAE} & This solution implements a pre-trained strategy inspired by Masked Auto-Encoders (MAEs). In addition to visualisations, tutorials are presented. The solution is implemented in Pytorch.\\
    \bottomrule
  \end{tabular}
\end{table*}

\cref{tab.3} presents results obtained by MTPSL \cite{MTPSL} on two commonly used MTL datasets: NYUv2 \cite{NYUv2} and Cityscapes \cite{cityscapes}. The results are reported on three tasks for NYUv2 \cite{NYUv2} including semantic segmentation, depth estimation and surface normals. Additionally, the results are reported on semantic segmentation and depth estimation for Cityscapes \cite{cityscapes}. 
First, MTPSL \cite{MTPSL} evaluates its cross-task consistency mapping method under two data availability settings. The first configuration consists of $\frac{1}{3}$ of the images, labelled with the three tasks, noted as MTPSL (1/3). The results reported in this setting suggest a degradation in performance compared to the single task learning (STL) baselines. However, the other setting, consisting of all images being labelled with only one of the tasks and noted as MTPSL (one) present better results closer to the STL baseline for all tasks. Although the two data settings present the same labeling demand, they showcase different performance. Therefore, this difference demonstrates that the joint space mapping is efficient \cite{MTPSL} under semi-supervised settings. 
Moreover, MTPSL \cite{MTPSL} displays, as part of their evaluation on Cityscapes \cite{cityscapes}, that some tasks are worth being shared more than others. The authors introduce an imbalanced supervision paradigm option and choose to use only 10\% of a task whilst keeping 90\% of the other task, noted as MTPSL (1:9), meaning 10\% of input images are annotated with segmentation ground truth and 90\% are labelled with depth ground truth. The results for imbalanced tasks present strong robustness, whereas the advantaged tasks outperform STL baselines. 

Similarly, \cref{tab.4} reviews results obtained by MTPSL \cite{MTPSL} on the Pascal-Context \cite{PASCAL} dataset under the 'one' data availability setting (where only one task label is available) for traditionally approached dense prediction tasks. We notice the major superiority of MTL under this setting : whilst still performing 5 tasks, \cite{MTPSL} manages to outperforms STL baselines on semantic segmentation and edge detection and still perform similarly to STL baselines on other tasks. 

\cref{tab.6} shows a range of publicly available code repositories for MTL including paper repositories, programming framework, benchmarking and partially-supervised code resources.

\section{Conclusion}
This review provided an extensive and comprehensive analysis of MTL systems in Computer Vision. 
Firstly, this work studied how architectural implications impact parameter sharing across tasks. 
Second, we analysed the concept of negative transfer and introduced MTL methods to remedy this issue through balancing the pace to which tasks learn during the training of a MTL system.
Third, this paper briefly reviewed how task relationships can be leveraged to provide new insights to task hierarchies to further improve the performance of MTL systems.
Fourth, we extensively reviewed how MTL can be utilised under partially supervised settings, for instance, as a self-supervised pre-training strategy for representation learning, or by exploiting task relationships to reduce the demand for labelled tasks in semi-supervised learning or finally by enhancing few-shot target tasks through cross-task parameter sharing.
Last, we summarised common multi-task datasets and code repositories to provide the interested reader with the necessary toolkits. 
We provide an analysis of results for partially-supervised MTL techniques. Our key insights for future work under this paradigm are: (1) MTL generally processes a small and constrained set of presumably related tasks. We identify there is a lack of adaptive methods, capable of learning relevant features from a large pool of tasks; otherwise, (2) reported results suggest partially-supervised MTL can be as performant as its fully-supervised single-task counterparts, sometimes even better whilst still providing output for multiple tasks : see \cref{tab.3}, \cref{tab.4} and \cref{tab.5} (\ie \textit{Few-Shot Learning, Semi-Supervised Learning}). There is therefore a need to explore solutions and data availability constraints under a multi-task framework. Finally, (3) we identify that MTL requires more benchmarking tools on large datasets. Taskonomy \cite{taskonomy} is the first step towards this direction and similar work could bring new insights to future research in MTL. 

\noindent \textbf{Acknowledgments.}
The authors would like to thank Prof. Tomasz Radzik for helpful discussions and  acknowledge the use of the King’s Computational Research, Engineering and Technology Environment (CREATE). Miaojing Shi was supported by the Fundamental Research Funds for the Central Universities.


%





\ifCLASSOPTIONcaptionsoff
  \newpage
\fi





\bibliographystyle{IEEEtranN}

%

\begin{IEEEbiography}[{\includegraphics[width=1in,height=1.25in,clip,keepaspectratio]{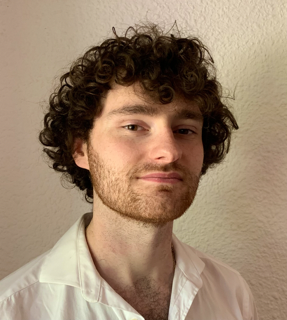}}]{Maxime Fontana}
received his BSc and his MSc in Computer Science from the University of Sheffield. He is currently a Ph.D. candidate from King's College London, United Kingdom. His research interests include computer vision, machine learning, real-time rendering, scene understanding and autonomous driving. His current research involves the development of innovative, more data-efficient multi-task learning systems.
\end{IEEEbiography}
\begin{IEEEbiography}[{\includegraphics[width=1in,height=1.25in,clip,keepaspectratio]{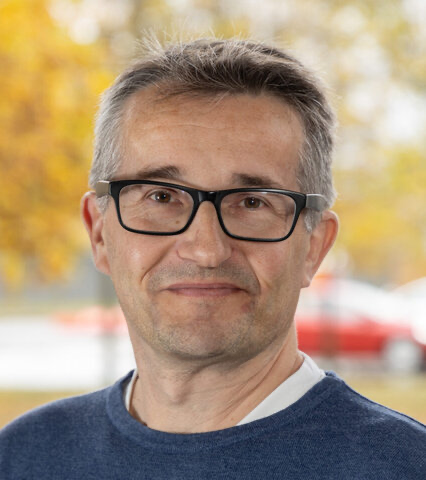}}]{Michael Spratling}'s research is concerned with understanding the computational and neural mechanisms underlying visual perception, and developing biologically-inspired neural networks to solve problems in computer vision and machine learning. He has a multidisciplinary background having trained and held posts in engineering, psychology, and computer science at various universities (Loughborough, Edinburgh, St Andrews, Cambridge, Birkbeck, and King's College London). He is currently a researcher in the Department of Behavioural and Cognitive Sciences at the University of Luxembourg.
\end{IEEEbiography}
\begin{IEEEbiography}[{\includegraphics[width=1in,height=1.25in,clip,keepaspectratio]{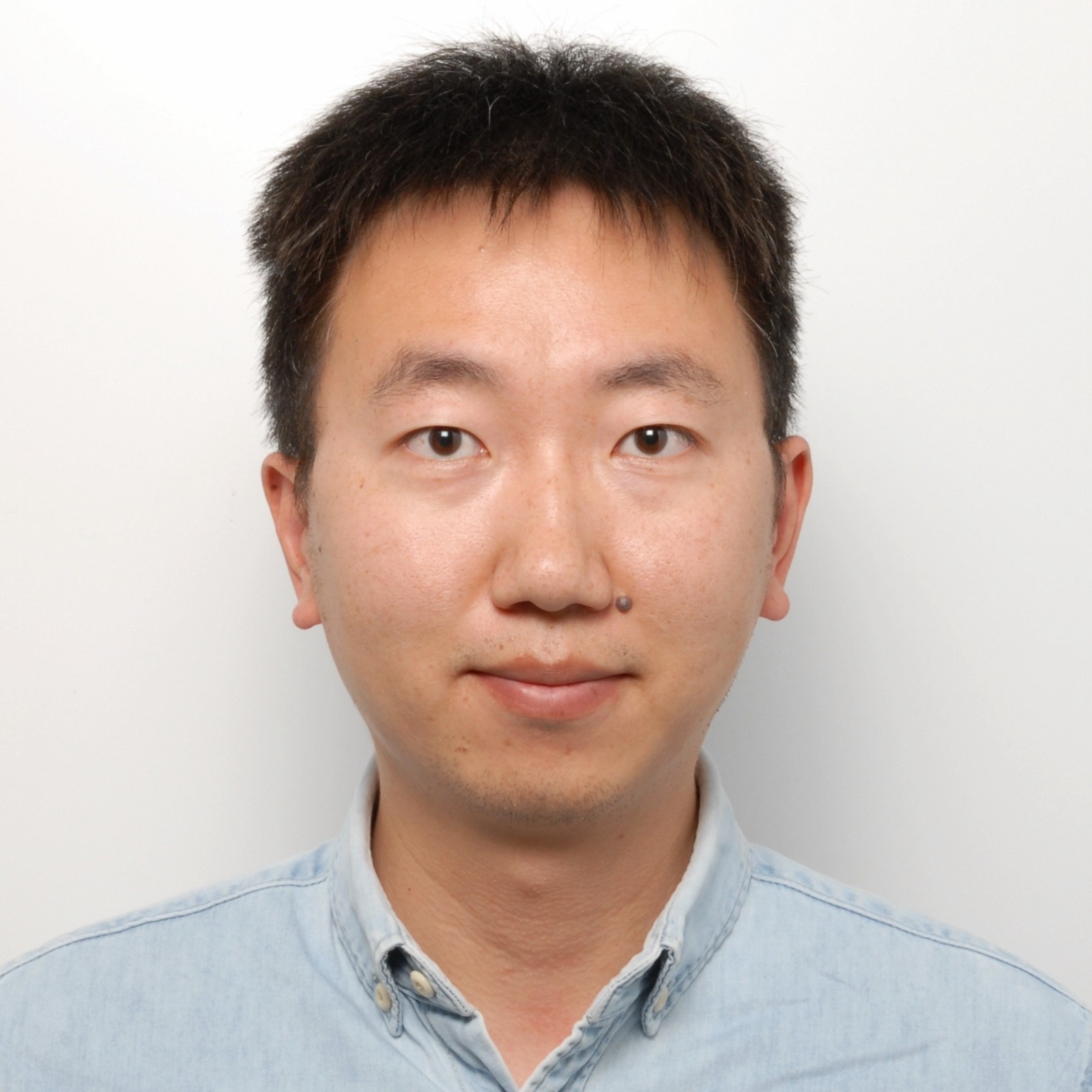}}]{Miaojing Shi}
(Senior Member, IEEE) received the Ph.D. degree from Peking University in 2015. He also engaged with a joint Ph.D. program with the University of Oxford and INRIA Rennes for a year. He held a postdoctoral position at the University of Edinburgh and was a Research Scientist at INRIA Rennes. Between 2020 and 2022, he has been a Lecturer/Senior Lectuer with the Department of Informatics, King's College London. Since 2023, he becomes a Full Professor at Tongji University and a visiting Senior Lecturer at King's. He has authored or co-authored over 70 papers in prestigious journals such as IEEE Transactions on Pattern Analysis and Machine Intelligence and Proceedings of the IEEE, as well as top AI conferences including CVPR, ICCV, NeurIPS, among others. His current research focus is on visual learning with few data, vision-language learning and medical imaging analysis.
\end{IEEEbiography}





\vfill


\end{document}